\documentclass{article}
\usepackage{arxiv}

\usepackage[utf8]{inputenc}
\usepackage[T1]{fontenc}
\usepackage{hyperref}
\usepackage{url}
\usepackage{booktabs}
\usepackage{amsfonts}
\usepackage{nicefrac}
\usepackage{microtype}
\usepackage{graphicx}
\usepackage{amsmath}
\usepackage{natbib}
\usepackage{doi}
\usepackage{xcolor}
\usepackage{array}
\newcommand{\samethanks}[1][\value{footnote}]{\footnotemark[#1]}

\title{Can VLMs Reliably Assess Sidewalk Accessibility Attributes from Pedestrian-Level Imagery?}

\author{ \href{https://orcid.org/0000-0003-3512-0822}{\includegraphics[scale=0.06]{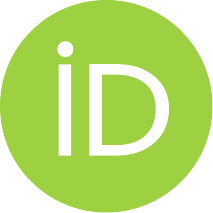}\hspace{1mm}Seung Jae Lieu}\thanks{These authors contributed equally.}\\
	Senseable City Lab\\
	Massachusetts Institute of Technology\\
	Cambridge, MA 02139 \\
	\texttt{lsj97@mit.edu} \\
    \And
\href{https://orcid.org/0000-0001-7275-6219}{\includegraphics[scale=0.06]{orcid.pdf}\hspace{1mm}Diego Morra}\samethanks\\
	Senseable City Lab\\
	Massachusetts Institute of Technology\\
	Cambridge, MA 02139 \\
	\texttt{d\_morra@mit.edu} \\
    \And
\href{https://orcid.org/0009-0004-6845-7987}{\includegraphics[scale=0.06]{orcid.pdf}\hspace{1mm}Chiara Cadoni   }\\
	Senseable City Lab\\
	Massachusetts Institute of Technology\\
	Cambridge, MA 02139 \\
	\texttt{ccadoni@mit.edu} \\
    \And
\href{https://orcid.org/0009-0005-6981-148X}{\includegraphics[scale=0.06]{orcid.pdf}\hspace{1mm}Wonseop Song}\\
	Senseable City Lab\\
	Massachusetts Institute of Technology\\
	Cambridge, MA 02139 \\
	\texttt{wonseop@mit.edu} \\
    \And
\href{https://orcid.org/0000-0001-8777-3927}{\includegraphics[scale=0.06]{orcid.pdf}\hspace{1mm}Martina Mazzarello}\\
	Senseable City Lab\\
	Massachusetts Institute of Technology\\
	Cambridge, MA 02139 \\
	\texttt{mmazz@mit.edu} \\
    \And
    {Carlo Ratti} \\
	Senseable City Lab\\
	Massachusetts Institute of Technology\\
	Cambridge, MA 02139 \\
	\texttt{ratti@mit.edu} \\
    }    

\renewcommand{\shorttitle}{Can VLMs Reliably Assess Sidewalk Accessibility Attributes?}

\hypersetup{
pdftitle={Can VLMs Reliably Assess Sidewalk Accessibility Attributes from Pedestrian-Level Imagery?},
pdfsubject={cs.CV},
pdfauthor={Seung Jae Lieu, Diego Morra, Chiara Cadoni, Wonseop Song, Martina Mazzarello, Carlo Ratti},
pdfkeywords={Sidewalk Accessibility, Vision-Language Model (VLM), Uncertainty Quantification, Conformal Prediction},
}

\begin{document}
\maketitle
\setcounter{footnote}{0}
\begin{abstract}

An important component of urban accessibility, particularly for wheelchair users and people with reduced mobility, is sidewalk compliance with measurable requirements. We test whether effective width, longitudinal slope, cross slope, and pavement condition can be assessed reliably from pedestrian-level imagery using vision-language models (VLMs). We present the first application of sampling-based conformal prediction (CP) for VLM-based accessibility assessment. We evaluate four VLMs on 514 sidewalk images from Seoul, South Korea, with field-measured ground truth. Conformal calibration attains the nominal 90\% coverage for all models and attributes, but the calibrated regions differ in informativeness. Effective width yields the most informative estimates, with a mean interval half-width of about 1.0\,m for the best model. Since every model overestimates width, asymmetric calibration shortens the intervals by up to 33\% at unchanged coverage. Longitudinal slope is marginally informative, cross-slope intervals are too wide to resolve regulatory thresholds, and pavement-condition sets degenerate to all five grades (A--E) for three of the four models. Uncalibrated intervals from raw sampling dispersion cover only 17--47\% of field-measured values at a nominal 90\% level. Among the images with the most self-consistent responses, these intervals miss the field-measured value in up to 96\% of cases. Response self-consistency is therefore not evidence of accuracy, and sampling dispersion cannot be interpreted as uncertainty until it has been calibrated against field-measured ground truth. No quantitative attribute reaches the precision required for general compliance assessment, but CP identifies from calibration data alone which attributes can support screening of segments far from the thresholds. We release the annotated pedestrian-level images and their corresponding field-measured attribute values.
\end{abstract}

\keywords{Sidewalk Accessibility \and Vision-Language Model (VLM) \and  Uncertainty Quantification \and Conformal Prediction}

\section{Introduction} \label{sec:intro}
Sidewalks provide access to services, public transportation, and civic life. Their accessibility benefits all pedestrians but is particularly critical for people with disabilities or with reduced mobility, such as older adults, for whom missing curb ramps, uneven surfaces, or other barriers can substantially limit independent mobility. In many jurisdictions, these principles are codified in accessibility regulations that define measurable design requirements \citep{ada2010standards, toronto2021accessibility, usaccessboard2023prowag, korea2025mobility} and assign responsibility for sidewalk provision, maintenance, and periodic compliance assessment to municipal governments \citep{Eisenberg2020barrier}. Despite these legal obligations, implementation remains limited where it has been assessed. In the United States, where implementation has been studied most extensively, a systematic evaluation of local governments showed that only 13\% published an Americans with Disabilities Act (ADA) transition plan, and among the jurisdictions reporting infrastructure conditions, 65\% of curb ramps and 48\% of sidewalks were classified as inaccessible \citep{Eisenberg2020barrier}. Transportation practitioners similarly identify the lack of comprehensive sidewalk inventories and scalable assessment methods as major barriers to accessibility planning and maintenance \citep{Wagner2025ADA}. These gaps carry practical and financial consequences: because the obligations are legally binding, non-compliance is a latent liability wherever they apply, and in the United States it has already materialized as litigation against municipalities \citep{eisenberg2024ADAmetrics}.
Yet comprehensive and up-to-date sidewalk accessibility data remain unavailable for most cities \citep{Froehlich2019accessible}. Existing assessment approaches rely primarily on built environment measures, which provide detailed measurements but require considerable time and resources, limiting update frequency \citep{browson2009measuring, ferrerfont2026scalablesidewalk}. Consequently, the pace at which accessibility conditions change often exceeds municipalities’ ability to monitor and address them, creating a persistent gap between what regulations require and what municipalities know about the current condition of sidewalks.

The increasing adoption of participatory governance mechanisms, together with the digitalization of public sector services, has led cities to develop platforms that facilitate direct communication with residents \citep{falco2019digital, helbing2024co}. These include, for example, the digitalization of non-emergency helplines in the form of online portals and smartphone applications\footnote{See for example \href{https://www.boston.gov/departments/boston-311}{Boston 311} or \href{https://play.google.com/store/apps/details?id=kr.go.seoul.mydata}{Seoul Helping Map}.} through which citizens can create and submit non-emergency service requests \citep{stowers2022city311}. These generally cover a range of urban issues such as trash or dead animal removal, pothole or general street maintenance, and graffiti removal, which citizens can report through a structured form and, in many cases, with photographic evidence \citep{Hartmann2017potential311}. Unlike periodic professional audits, or street-view imagery for which data currency has been identified as a practical concern by municipal stakeholders \citep{saha2019projectsidewalk}, these platforms provide municipalities with continuously updated and geographically distributed observations of infrastructure conditions. However, a recent study that analyzed more than seven million service requests from 29 US cities found that resolution times vary markedly across service categories, with evidence that safety and health-related requests are prioritized, while street and sidewalk repair requests exhibit the longest resolution times \citep{stowers2022city311}. Similar to what happens with road-damage assessment \citep{arya2021roaddamage}, the observed resolution delays associated with sidewalk maintenance requests could reflect the effort required to verify each report, assess its severity, and allocate limited resources accordingly, a burden that grows with the volume and the spatio-temporal variability of incoming requests \citep{xu2017predict311, mclafferty2020bias311}.

Automating this evaluation stage could substantially reduce the burden on city administration by screening reports for validity, severity, and accessibility implications before expert inspection. Existing computer vision approaches have demonstrated promising performance for sidewalk assessment, but most rely on specialized sensing equipment, expert-operated acquisition pipelines, or models trained for predefined accessibility attributes \citep{lee2026lidarSidewalks,hang2025neuralradiance}. Vision-language models (VLMs) have meanwhile emerged as a versatile tool for image-based assessment of the built environment \citep{Jang02012025}, scoring walkability and streetscape quality from street-level imagery \citep{blecic2024urbanwalkability,cai2025LLMurban,perez2025generativeAI}, and supporting accessibility assessment and annotation in contexts ranging from wheelchair mobility to data-scarce cities \citep{wang2026vlmssensorsfeelscalable,lalwani2026VLMannotation}. These studies suggest that VLMs could support image-based assessment of specific built-environment characteristics, including walkability and sidewalk accessibility. Unlike conventional computer vision systems, a single VLM can evaluate heterogeneous accessibility criteria expressed directly in natural language, performing multiple assessment tasks without task-specific retraining. These capabilities make VLMs potentially useful for municipal accessibility screening, where submitted imagery must be interpreted against regulatory criteria to identify invalid reports, estimate issue severity, and prioritize cases requiring expert inspection.

However, deploying VLMs in this role requires reliable estimates of prediction uncertainty, a capability that current models lack. This limitation is intrinsic to the stochastic nature of VLMs. Their outputs can vary across repeated queries and prompt formulations, and their performance is often heterogeneous across visual attributes and scene conditions. Moreover, model-reported uncertainty is frequently poorly aligned with empirical accuracy \citep{groot2024overconfidence, zhang2024vluncertaintydetecting}. In a municipal triage pipeline, these errors would have asymmetric operational consequences. False positives may direct inspection resources toward reports that do not require intervention \citep{mclafferty2020bias311}, whereas false negatives may cause valid accessibility issues to be deprioritized or dismissed, with potential compliance and liability implications.
In both cases, subsequent verification and correction reintroduce effort that automation is intended to reduce.
Operational deployment therefore requires more than a single estimated value; it requires an interval (or, for categorical attributes, a set of categories) whose probability of containing the true value is known. Here, we call a VLM assessment reliable when it meets two conditions: (i) the interval or set contains the field-measured value at a stated rate, for example in 90\% of cases, and (ii) the interval or set is narrow enough to determine whether the sidewalk meets the relevant regulatory threshold.
Existing approaches to uncertainty estimation, however, are poorly suited to this setting. Verbalized confidence is often miscalibrated and systematically overconfident \citep{xiong2024LLMUuncertainty}, while post-hoc calibration methods generally require access to internal model probabilities that proprietary API-based VLMs do not expose.

In this context, conformal prediction (CP) provides a framework for quantifying predictive uncertainty \citep{vovk2005algorithmic,angelopoulos2023conformalprediction}. Given a calibration set with ground truth annotations, CP transforms the output of an arbitrary predictive model into prediction sets or intervals that achieve a user-specified marginal coverage guarantee under the assumption of exchangeable data. 
These properties make CP particularly well suited to proprietary VLMs, which do not expose internal probabilities, so that uncertainty must be estimated from the responses the model returns.
Because this guarantee holds without requiring distribution-specific modeling or model retraining, CP has been employed in risk-sensitive settings ranging from clinical decision support to safe robotic planning \citep{angelopoulos2023conformalprediction}.
More recently, CP has been extended to VLMs, demonstrating its potential for uncertainty-aware multimodal reasoning \citep{su2024APIconformal}.

However, the two lines of work have not been combined. Existing evaluations of VLMs for urban accessibility assessment rely on aggregate metrics such as agreement rates and benchmark scores \citep{Dai2024systematicreview}, which describe average performance but do not indicate how much any individual prediction can be trusted.
To the best of our knowledge, sampling-based CP has not yet been applied to the assessment of the built environment or urban accessibility. \textbf{This raises the question of whether conformal prediction can provide statistically valid uncertainty estimates for VLM-based sidewalk assessment, and whether the resulting prediction intervals and sets are narrow enough to determine whether a sidewalk meets the regulatory threshold for each attribute.}
We investigate this question in a controlled setting designed to evaluate VLM-based sidewalk accessibility screening. Specifically, we evaluate four VLMs on a dataset of 514 pedestrian-level sidewalk images collected in Seoul, South Korea, with ground-truth annotations derived from expert field observations and measurements according to Korean accessibility regulations. In this study, we consider citizen-report triage for municipal accessibility screening as a motivating use case rather than the setting evaluated: the images were collected using a standardized acquisition protocol and do not represent citizen submissions. This controlled setting allows us to establish which attributes can be assessed reliably before extending the evaluation to more heterogeneous citizen-generated imagery.
Our contributions are fourfold:
\begin{enumerate}
    \item The first application of CP to VLM-based sidewalk accessibility assessment, evaluating its suitability for statistically calibrated assessment of sidewalk attributes defined in accessibility regulations.
    \item A systematic evaluation of two proprietary API-based and two open-weight VLMs, using sampling-based CP, which relies on the variation among repeated model responses rather than on internal token probabilities.
    \item An attribute-by-attribute analysis of CP for three continuous attributes, effective width, longitudinal slope, and cross slope, for which CP returns intervals, and one ordered attribute, pavement condition graded A--E, for which CP returns sets of grades.
    \item The release of a benchmark dataset of 514 pedestrian-level sidewalk images with expert field annotations.
\end{enumerate}

\section{Related work} \label{sec:related}

\subsection{AI-Based Sidewalk Accessibility Assessment}
\label{sec:AIbased-assessment}

Research on vision-based sidewalk accessibility assessment has evolved from task-specific computer vision models toward foundation models capable of evaluating heterogeneous accessibility attributes.
Early work focused on automating accessibility assessment using task-specific computer vision models.
Deep-learning detectors have been developed to validate and generate accessibility labels from street-view imagery \citep{weld2019sidewalkstreetscape}, trained on annotations collected through large-scale crowdsourcing platforms \citep{saha2019projectsidewalk}.
Subsequent research has investigated annotation quality and cross-city transferability \citep{duan2022crowdAI} and introduced increasingly fine-grained accessibility benchmarks \citep{liu2024finegrainedsidewalk}. Similar advances have extended to large-scale semantic segmentation, enabling extraction of sidewalk networks from aerial imagery \citep{hosseini2023aerial}.

When geometric accuracy is required, richer sensing technologies become the dominant approach. Mobile LiDAR enables network-scale inventories of sidewalk width, grade, and cross slope for accessibility compliance assessment \citep{hou2020networklevel}, while neural radiance fields (NeRF) reconstruct audit-grade geometry from consumer-camera videos \citep{hang2025neuralradiance}. More recently, dedicated geometric estimation pipelines have been proposed for sidewalk measurements from street-view imagery \citep{tan2026urbanvggt}. These approaches substantially improve measurement fidelity but require dedicated sensing hardware, specialized acquisition procedures, or computationally intensive processing. Conversely, conventional street-view imagery is inexpensive and readily available, yet it provides limited refresh frequency and is captured from the roadway, systematically misrepresenting the pedestrian environment \citep{ki2023humancentric, ito2024streetview}.
Collectively, these approaches illustrate a recurring trade-off between measurement fidelity and deployment scalability. Specialized geometric pipelines can achieve relatively low measurement error under controlled validation, but their cross-city deployment remains constrained by image availability, calibration assumptions, filtering, and the lack of local ground truth \citep{tan2026urbanvggt}.

These computer vision pipelines remain specialized to predefined accessibility attributes, requiring dedicated labels, training procedures, or sensing workflows for each assessment task. VLMs fundamentally change this paradigm by enabling heterogeneous accessibility criteria to be evaluated through natural-language instructions using a single pretrained model. Recent work has therefore begun exploring their application to urban scene understanding and accessibility assessment \citep{peng2025vision}. Multimodal VLM-based audits of the built environment have demonstrated strong agreement with established virtual-audit methodologies \citep{Jang02012025}.
VLMs have been used to jointly estimate walkability scores and generate qualitative explanations \citep{blecic2024urbanwalkability}, and encoding expert assessment protocols within prompts has been shown to concentrate the resulting score distributions \citep{cai2025LLMurban}.
At the same time, generative streetscape scoring reports its weakest accuracy precisely on metric attributes such as sidewalk width \citep{perez2025generativeAI}. Within accessibility assessment, expert-guided VLM evaluations have shown moderate aggregate agreement with sensor-derived wheelchair mobility signals \citep{wang2026vlmssensorsfeelscalable}, and VLM-assisted annotation has supported accessibility mapping in data-scarce environments \citep{lalwani2026VLMannotation}.

Across both generations of methods, however, evaluation remains dominated by aggregate performance metrics, including agreement rates, inter-rater reliability, and model benchmark scores \citep{Dai2024systematicreview}. While these measures quantify average model performance, they provide no information about how often an individual prediction contains the true value. This limitation is particularly important for accessibility auditing, where individual assessments, not average benchmark performance, support inspection, maintenance prioritization, and regulatory compliance decisions. VLMs therefore offer a potentially scalable approach to accessibility assessment, but their outputs remain point predictions without the formal reliability guarantees required for compliance-oriented auditing.

\subsection{Reliability and Calibration of VLM Outputs}
\label{sec:related-reliability}

Beyond the lack of statistical guarantees discussed in the previous section, VLM predictions are also inherently unstable. Model outputs may vary across repeated queries even under nominally deterministic decoding \citep{atil2025nondeterminism}, and can change substantially under semantically irrelevant modifications to prompt formatting, leading to accuracy variations \citep{sclar2024quantifyllm}. In an accessibility auditing pipeline, this implies that the same image evaluated under the same assessment criterion may yield different predictions across repeated executions. Such variability represents an additional source of uncertainty that should be explicitly quantified.

A natural response is to ask the model to estimate its own confidence, an approach that can improve on the model's own conditional probabilities \citep{tian2023justask}.
However, verbalized confidence has repeatedly been shown to be systematically miscalibrated and overconfident for large language models \citep{xiong2024LLMUuncertainty}, and the same holds for VLMs, which report high confidence at chance-level accuracy and can produce means, standard deviations, and confidence intervals that are almost uninformative about their own error \citep{groot2024overconfidence}.
A second family of approaches addresses uncertainty through post-hoc calibration of model probabilities. Techniques such as temperature scaling adjust predicted probabilities using held-out calibration data \citep{guo2017calibration}, and a broad range of calibration methods has subsequently been developed for language models \citep{geng2024survey}. These approaches, however, require access to internal probability distributions or token-level logits, which are generally unavailable for proprietary VLM APIs.

Under black-box access, uncertainty estimation relies primarily on repeated sampling of model outputs, scored by the dispersion of the resulting responses. Self-consistency aggregates them by majority vote \citep{wang2023selfconsistency}, whereas semantic-entropy approaches cluster them by meaning and score the entropy of that distribution. The latter have been extended to multimodal models by perturbing both image and question in semantically equivalent ways \citep{zhang2024vluncertaintydetecting}, within a rapidly expanding methodological landscape \citep{shorinwa2025survey}.
Although these approaches often correlate with prediction reliability, they remain heuristic and are not designed to provide statistically valid coverage guarantees for individual predictions. Moreover, empirical studies show that these uncertainty estimates are often poorly calibrated \citep{savage2025large}, while benchmark evaluations of VLMs similarly report weak correspondence between estimated uncertainty and actual prediction accuracy \citep{kostumov2024uncertaintyaware}. More fundamentally, dispersion-based methods implicitly assume that variability across repeated responses is informative of prediction error. This assumption does not always hold, as models may repeatedly generate the same incorrect prediction, and such self-consistent errors have been shown to persist, and in some cases become more frequent, as model capability increases \citep{tan2025consistent}. Consequently, response consistency alone cannot be interpreted as evidence of prediction correctness.

Existing approaches face complementary limitations. Calibration methods with formal statistical guarantees require access to model internals that commercial APIs do not expose, whereas sampling-based heuristics remain compatible with black-box models but primarily optimize confidence estimation rather than statistically valid coverage. CP addresses this gap by providing finite-sample coverage guarantees without requiring access to model internals. Importantly, it does not eliminate prediction instability. It instead quantifies uncertainty in the presence of that instability by constructing prediction sets whose long-run coverage can be formally controlled under the assumption of exchangeable calibration and deployment data.

\subsection{Conformal Prediction for Language and Vision-Language Models}
\label{sec:related-cp}

CP is a general framework for uncertainty quantification that transforms the output of an arbitrary predictive model into prediction sets or intervals with finite-sample coverage guarantees under the assumption of exchangeable data \citep{vovk2005algorithmic, shafer2008tutorial, angelopoulos2023conformalprediction}. The split (inductive) formulation requires only a calibration set and a single calibration phase \citep{papadopoulos2002inductiveconfidence}, making it practical for large-scale applications. Mature conformal methods are available for both continuous and categorical prediction tasks, including conformalized quantile regression (CQR) for continuous variables \citep{romano2019quantile} and least-ambiguous class (LAC) and adaptive prediction sets (APS) for classification \citep{Sadinle2019classifier, romano2020neural}. Recent surveys document the rapid adoption of CP across machine learning, natural language processing, and data-centric AI \citep{campos2024conforlamNL, zhou2025data}.

Recent work has extended CP to foundation models. For language models, conformal methods have been proposed for calibrated generation \citep{quach2024conformal}, factuality-aware generation \citep{mohri2024languagemodels}, and conditional validity \citep{cherian2024LLMvalidity}. Similar developments have emerged for VLMs, including zero-shot image classification \citep{Silva2025CVPR}, medical image analysis \citep{silva2026fullconformal}, visual question answering \citep{ye2025datadrivencalibration}, radiology report generation \citep{alyassirad2026conrep}, learned abstention policies \citep{tayebati2025conformalpolicies}, and broader evaluations of foundation models as conformal predictors \citep{fillioux2026foundationmodels}.

Particularly relevant are recent sampling-based conformal approaches, which replace inaccessible logits with statistics derived from repeated model queries. Proposed nonconformity measures include response frequency combined with normalized entropy and semantic similarity \citep{su2024APIconformal} and frequency-based predictive entropy \citep{yang2025frequency, yang2025conformalsets}. Complementary work addresses exchangeability violations through conformal outlier screening \citep{wang2025sconu} and calibration reweighting under domain shift \citep{lin2026domainshift}, alongside empirical-Bayes uncertainty estimation \citep{zeng2026empirical}.

Despite this growing body of work, important gaps remain. First, to the best of our knowledge, sampling-based CP has not been investigated for sidewalk accessibility assessment. Correspondingly, none of the accessibility assessment pipelines reviewed in Section~\ref{sec:AIbased-assessment} provides statistically valid uncertainty estimates for individual predictions.

Second, existing sampling-based conformal methods have been developed and evaluated primarily on language and multimodal reasoning benchmarks involving discrete textual outputs. Sidewalk accessibility auditing represents a substantially different evaluation setting, as it combines heterogeneous prediction types (continuous and ordered discrete attributes), predictions interpreted against explicit regulatory standards, and validation through field-measured ground truth. Finally, repeated VLM sampling may produce highly consistent but incorrect predictions, a regime that has received little empirical attention in existing sampling-based CP research.

\section{Methodology} \label{sec:methodology}

\begin{figure}[t]
    \centering
    \includegraphics[width=\linewidth]{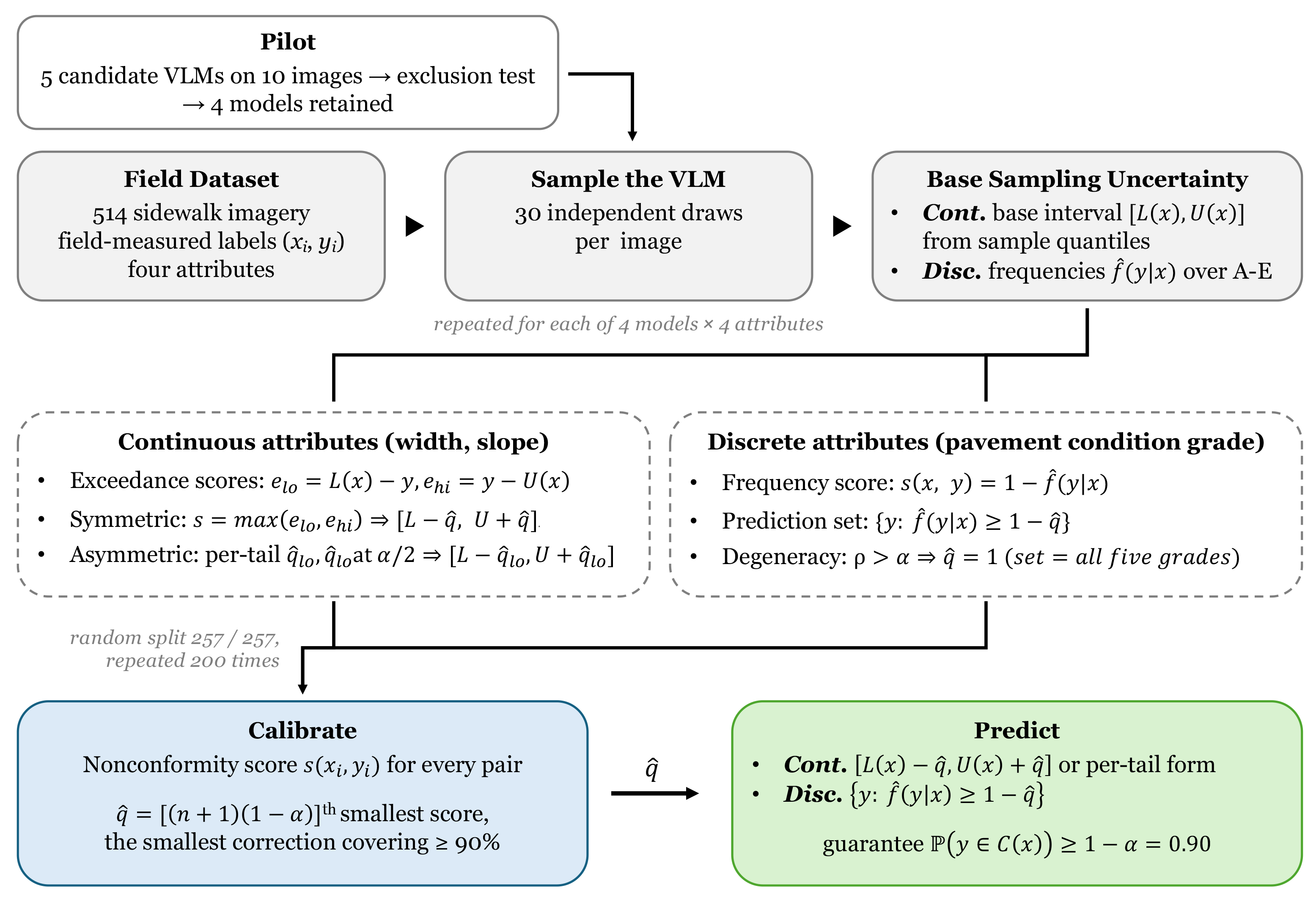}
    \caption{Overview of the sampling-based CP pipeline. A pilot study on ten images preceded the full evaluation and led to the exclusion of the Claude model, whose responses showed no sampling variation.}
    \label{fig:fig_1}
\end{figure}

Figure~\ref{fig:fig_1} summarizes the entire pipeline: field data collection,
repeated sampling of each VLM, construction of base uncertainty representations,
split conformal calibration, and the criteria against which the calibrated
regions are evaluated. The remainder of this section details each component.

\subsection{Conformal prediction framework}
\label{sec:cp-framework}

The uncertainty quantification procedure in this study is an application of split CP \citep{papadopoulos2002inductiveconfidence, vovk2005algorithmic}, in the sampling-based formulation of \citet{su2024APIconformal}. The methodology is established, and the reader is referred to \citet{angelopoulos2023conformalprediction} for the underlying theory. This section records the operational procedure and the shared notation, while Sections~\ref{sec:continuous} and~\ref{sec:discrete} give the attribute-specific instantiations. The sampling-based formulation is dictated by the model set itself. Our evaluation spans proprietary API models, which expose neither token-level logits nor internal representations, and open-weight models, which do. Logit-based nonconformity scores are unavailable for the former, and adopting them for the latter would entangle differences between models with differences between scoring procedures. We instead query every model through the same black-box interface and build every score from the same observable statistics over repeated responses, so that a single procedure applies uniformly across the model set and any performance difference reflects the models.

Operationally, the framework involves four steps.
\begin{enumerate}
    \item \emph{Sampling.} For each image $x$ and attribute, the model is queried $M = 30$ times, retaining the parsed responses $\{\hat{y}^{(1)}(x), \ldots, \hat{y}^{(M)}(x)\}$.
    \item \emph{Scoring.} A nonconformity score $s(x, y) \in \mathbb{R}$ measures how poorly a candidate value $y$ agrees with the sampled responses for $x$. This score is the only attribute-specific ingredient of the framework (Sections~\ref{sec:continuous} and~\ref{sec:discrete}).
    \item \emph{Calibration.} Given a calibration set $\mathcal{D}_{\mathrm{cal}} = \{(x_i, y_i)\}_{i=1}^{n}$ of images paired with field-measured ground truth, disjoint from the test set, the calibrated quantile at a user-specified risk level $\alpha$ is
    \begin{equation}
        \hat{q} \;=\; \mathrm{Quantile}\Big( \{ s(x_i, y_i) \}_{i=1}^{n},\;
        \tfrac{\lceil (n+1)(1-\alpha) \rceil}{n} \Big).
    \label{eq:quantile}
    \end{equation}
    \item \emph{Prediction.} For a new image, the region collects every candidate value whose score clears the calibrated threshold,
    \begin{equation}
        \mathcal{C}_\alpha(x_{\mathrm{test}})
        \;=\; \{\, y \mid s(x_{\mathrm{test}}, y) \le \hat{q} \,\}.
    \label{eq:region}
    \end{equation}
\end{enumerate}

Under exchangeability of calibration and test data, this construction satisfies the finite-sample guarantee

\begin{equation}
    \mathbb{P}\big( y_{\mathrm{test}} \in \mathcal{C}_\alpha(x_{\mathrm{test}})
    \big) \;\ge\; 1 - \alpha
\label{eq:coverage}
\end{equation}

for any model and any data distribution, without retraining \citep{angelopoulos2023conformalprediction}.
This guarantee has two important implications for interpreting the results. First, validity holds whether or not the model is accurate, as a poor score yields regions that are valid but wide. The empirical question throughout this study is therefore whether the calibrated regions are narrow enough to be operationally useful. Second, because our scores are heavily tied (frequencies take at most $M + 1$ values and continuous responses lie on a one-decimal grid), empirical coverage moderately above the nominal level is expected rather than anomalous \citep{romano2019quantile}.

The four attributes require two instantiations of this procedure. Sidewalk width, longitudinal slope, and cross slope are continuous, with intervals as output (Section~\ref{sec:continuous}). The pavement condition grade is ordinal on the five-level A--E rubric, with sets of candidate grades as output (Section~\ref{sec:discrete}). Both share Equations~(\ref{eq:quantile})--(\ref{eq:coverage}) and differ only in $s(x, y)$. Coverage is directly comparable across the full attribute set.

\subsubsection{Continuous attributes: conformalized sampling quantiles}
\label{sec:continuous}

For a continuous attribute, the $M$ responses form an empirical predictive distribution, summarized here by a central interval. Let $\hat{Q}_x(\cdot)$ denote the empirical quantile function of the samples for image $x$. The base interval is

\begin{equation}
    \big[ L(x),\, U(x) \big]
    \;=\;
    \Big[ \hat{Q}_x\big(\tfrac{\beta}{2}\big),\;
    \hat{Q}_x\big(1 - \tfrac{\beta}{2}\big) \Big].
\label{eq:base}
\end{equation}

Matching the base level $\beta$ to the conformal target $\alpha$ is a convention. The guarantee of Equations~(\ref{eq:quantile})--(\ref{eq:coverage}) holds for any base interval fixed before calibration, and the choice affects only the shape of the intervals, not their coverage.

Two properties of Equation~(\ref{eq:base}) matter in practice. First, with $M = 30$ the endpoints are extreme order statistics. At $\alpha = 0.1$ they are roughly the second smallest and second largest responses, the finest representable tail level is $1/M \approx 0.033$, and the base interval degenerates to the sample range whenever $\alpha < 2/M \approx 0.067$. The endpoints remain statistically noisy above that floor. Second, the interval is adaptive, contracting where repeated responses agree and widening where they disperse, but it carries no coverage guarantee, because variability under stochastic decoding is not the same as deviation from ground truth. Confidence intervals elicited from language models on numerical estimation tasks are systematically overconfident \citep{epstein2025llms}, and Section~\ref{sec:results} shows that sampled-quantile intervals behave the same way for sidewalk geometry.

We therefore conformalize the base interval following CQR, replacing the fitted quantile regressor with the model's own sampling quantiles. This substitution requires no logit access and extends the logit-free approach to continuous outputs. \citet{epstein2025llms} identify multi-sample empirical quantiles as one of two admissible routes to the base interval, but report conformal results only for direct prompting and leave the sampled route unevaluated on grounds of cost. Our pipeline supplies that empirical instantiation. 

Define the one-sided signed exceedances

\begin{equation}
    e^{\mathrm{lo}}_i = L(x_i) - y_i,
    \qquad
    e^{\mathrm{hi}}_i = y_i - U(x_i),
\label{eq:exceed}
\end{equation}

which are positive exactly when $y_i$ falls below or above the base interval. We consider two calibration variants.

\paragraph{Symmetric calibration}
The pooled score $s(x_i, y_i) = \max\{e^{\mathrm{lo}}_i, e^{\mathrm{hi}}_i\}$ records the signed distance by which the base interval fails to contain the truth, and is negative when $y_i$ lies inside. Substituting into Equation~(\ref{eq:quantile}) yields a single correction $\hat{q}$ and the interval

\begin{equation}
    \mathcal{C}_\alpha(x_{\mathrm{test}})
    \;=\;
    \big[ L(x_{\mathrm{test}}) - \hat{q},\;
    U(x_{\mathrm{test}}) + \hat{q} \big].
\label{eq:symmetric}
\end{equation}

A negative $\hat{q}$ is admissible and contracts conservative base intervals. Validity follows directly from Equations~(\ref{eq:quantile})--(\ref{eq:coverage}).

\paragraph{Asymmetric calibration}
Each endpoint is calibrated against its own tail. Fix an allocation $\alpha_{\mathrm{lo}} + \alpha_{\mathrm{hi}} = \alpha$ before seeing the calibration labels, here the equal split $\alpha_{\mathrm{lo}} = \alpha_{\mathrm{hi}} = \alpha/2$, and set

\begin{equation}
    \hat{q}_{\bullet}
    \;=\;
    \mathrm{Quantile}\Big( \{ e^{\bullet}_i \}_{i=1}^{n},\;
    \tfrac{\lceil (n+1)(1-\alpha_{\bullet}) \rceil}{n} \Big),
    \qquad \bullet \in \{\mathrm{lo}, \mathrm{hi}\},
\label{eq:asym-q}
\end{equation}

\begin{equation}
    \mathcal{C}_\alpha(x_{\mathrm{test}})
    \;=\;
    \big[ L(x_{\mathrm{test}}) - \hat{q}_{\mathrm{lo}},\;
    U(x_{\mathrm{test}}) + \hat{q}_{\mathrm{hi}} \big].
\label{eq:asymmetric}
\end{equation}

This variant runs two one-sided conformal procedures in parallel. Each correction controls its own tail at the assigned level, and a union bound over the two failure events gives $\mathbb{P}(y_{\mathrm{test}} \in C_\alpha(x_{\mathrm{test}})) \ge 1 - \alpha$, matching the two-tailed guarantee of \citet[Theorem~2]{romano2019quantile}. The guarantee holds for any allocation fixed independently of the calibration labels, hence the a priori equal split. Estimating two tail quantiles from the same $n$ observations carries greater estimation variance, an effect negligible at the calibration sizes used here.

The motivation for the asymmetric variant is specific to metric estimation from imagery. A model that systematically over- or under-estimates a physical dimension produces misses concentrated in one tail, and the pooled quantile of Equation~(\ref{eq:symmetric}) then displaces the endpoint on the well-behaved side further than coverage requires. Whether this occurs is an empirical property of each model and attribute pair, which we diagnose through the base-interval violation rates

\begin{equation}
    \pi_{\mathrm{lo}}
    = \tfrac{1}{n} \textstyle\sum_i \mathbb{1}\big[\, y_i < L(x_i) \,\big],
    \qquad
    \pi_{\mathrm{hi}}
    = \tfrac{1}{n} \textstyle\sum_i \mathbb{1}\big[\, y_i > U(x_i) \,\big],
\label{eq:violation}
\end{equation}

reported per attribute in Section~\ref{sec:results}. The answer should not be assumed to transfer between attributes. Width is a strictly positive magnitude, whereas the slopes are signed quantities distributed about zero, for which a magnitude bias would manifest symmetrically.

\subsubsection{Discrete attributes: frequency-based prediction sets}
\label{sec:discrete}

For the pavement condition grade, the response space is the ordered finite set $\mathcal{Y} = \{A, B, C, D, E\}$ and the output is a set of candidate grades. Without logit access, the softmax scores on which LAC \citep{Sadinle2019classifier} and APS \citep{romano2020neural} rely are unavailable; hence, we use the empirical sampling frequency in their place \citep{su2024APIconformal},

\begin{equation}
  \hat{f}(y \mid x) \;=\; \frac{1}{M}\sum_{m=1}^{M}
  \mathbb{1}\!\left[\hat{y}^{(m)}(x) = y\right].
\label{eq:freq}
\end{equation}

Its complement, $s(x, y) = 1 - \hat{f}(y \mid x)$, is the LAC score with the sampling frequency substituted for the softmax probability. The same score is adopted for black-box multiple-choice question answering by \citet{yang2025conformalsets}. Substituting into Equations~(\ref{eq:quantile})--(\ref{eq:region}), the prediction set collects every grade whose sampled frequency clears the calibrated threshold,

\begin{equation}
  \mathcal{C}_\alpha(x_{\mathrm{test}}) \;=\;
  \big\{\, y \in \mathcal{Y} : \hat{f}(y \mid x_{\mathrm{test}}) \ge 1 - \hat{q} \,\big\}.
\label{eq:lac-set}
\end{equation}

When the true grade of a calibration image never appears among its $M$ samples, Equation~(\ref{eq:freq}) is zero and $s(x, y)$ attains its maximum of $1$. We retain such cases at that maximal value, so that model failures are reflected in the calibration quantile rather than silently excluded. 

Two properties of this frequency-based construction, score concentration and degeneracy, shape its behavior. Both prove consequential in Section~\ref{sec:results}.

\paragraph{Score concentration} 
With $M = 30$ the attainable frequencies are multiples of $1/30$. Scores take at most $M + 1$ distinct values and ties are common, coarsening the achievable threshold and inflating set sizes \citep{su2024APIconformal}. Estimating the underlying output probabilities to finer resolution by additional sampling is prohibitively costly. \citet{su2024APIconformal} address the concentration by adding two fine-grained terms to the frequency, namely the normalized predictive entropy of the sampled distribution and the semantic similarity of each response to the modal response. The similarity term is vacuous for a five-letter response space, and no fine-grained refinement can lower the score of a grade that is never sampled, which is the regime characterized next. We therefore retain the frequency-only score. Since any measurable function of the observed samples is a valid nonconformity score, this choice does not affect the guarantee.

\paragraph{Degeneracy under systematic misgrading} Equation~(\ref{eq:lac-set}) is informative only if the true grade appears among the $M$ samples with non-trivial frequency. Let $\rho$ denote the proportion of calibration images whose true grade is never sampled; these images carry the maximal score. With $k = \lceil (n+1)(1-\alpha) \rceil$ the calibrated threshold satisfies $\hat{q} = 1$ whenever $k > (1 - \rho)\, n$, that is, asymptotically whenever

\begin{equation}
    \rho \;>\; \alpha .
\label{eq:degeneracy}
\end{equation}

Equation~(\ref{eq:lac-set}) then admits every grade for every image. Coverage is then trivially $100\%$ and the prediction conveys nothing. The procedure remains valid, in that it correctly reports that the model cannot resolve the attribute, but it is uninformative. 

\subsection{Evaluation criteria}
\label{sec:eval-criteria}
The guarantee of Equation~(\ref{eq:coverage}) makes validity a property of the procedure rather than of the model. Empirical coverage alone cannot distinguish a useful predictor from a useless one. We therefore evaluate each model and attribute pair against criteria that follow the output type. Continuous attributes are judged on their calibrated intervals and the discrete attribute on its calibrated grade sets.

\subsubsection{Criteria for continuous attributes}

\paragraph{Validity}
Empirical coverage of the calibrated regions on the test set, compared with the nominal level $1 - \alpha$ across the sweep $\alpha \in \{0.05, \ldots, 0.30\}$. Because the scores are heavily tied, coverage at or moderately above the nominal level is the expected signature of a correct implementation rather than evidence of conservatism.

\paragraph{Bias structure and calibration variants}
The one-sided violation rates $\pi_{\mathrm{lo}}$ and $\pi_{\mathrm{hi}}$ of Equation~(\ref{eq:violation}) diagnose systematic over- or under-estimation of each attribute. They determine whether the asymmetric variant of Section~\ref{sec:continuous} shortens the calibrated intervals relative to the symmetric one, and the comparison between the two variants is reported under this criterion.

\paragraph{Self-consistent-error diagnostic}
The adaptive base interval of Equation~(\ref{eq:base}) reads agreement across the $M$ sampled responses as low uncertainty. Whether that reading is warranted is tested by the base-interval violation rate within the narrowest quartile of base widths, which isolates the images on which the model is most self-consistent. This diagnostic is the continuous, per-image form of the self-consistent-error regime of \citet{tan2025consistent}, in which a model errs precisely where it is most assured. High violation rates in this quartile mean that response consistency cannot be read as confidence, and that coverage on confidently wrong images is preserved only through the uniform conformal correction, at a cost in width.

\subsubsection{Criteria for the discrete attribute}
\paragraph{Degeneracy}
We report $\rho$, the proportion of calibration images whose true grade is never sampled, together with the threshold condition of Equation~(\ref{eq:degeneracy}). When $\rho$ lies close to $\alpha$, individual calibration splits fall on either side of the condition, and averages over repeated partitions mix degenerate and informative splits into a summary that is realized on no single partition. We report the fraction of degenerate partitions alongside coverage and set size, and interpret per-split behavior separately in that regime.

\paragraph{Coverage and set size}
Empirical coverage of the calibrated grade sets is read against the nominal level exactly as for intervals, and the mean set size takes over the role of interval width as the measure of informativeness. A set of one grade is a sharp claim, while a set of all five grades conveys nothing, whatever its coverage.

\paragraph{Top-1 accuracy}
The share of images whose modal sampled grade equals the field grade. This quantity involves no calibration and characterizes the raw model. It matters because the frequency score of Equation~(\ref{eq:freq}) can only concentrate the set on grades the model actually samples. Low top-1 accuracy is thus the upstream cause of the failures that the remaining criteria detect.

\section{Study Design}\label{sec:study-design}
The experimental evaluation presented in this paper relies on a benchmark dataset designed to evaluate VLMs for sidewalk accessibility assessment. We constructed a new benchmark tailored to the accessibility dimensions defined by Korean regulations and paired each pedestrian-level image with manually acquired ground truth measurements. The following sections describe the adopted accessibility taxonomy, the dataset construction and anonymization procedures, and the selection of candidate VLMs used throughout the experimental evaluation.

\subsection{Sidewalk Attributes Taxonomy}\label{subsec:taxonomi}
To ground our attribute selection in Korean regulations, technical guidance, and field-assessment practice, we reviewed three principal sources, namely the Guidelines for Installation and Management of Sidewalks \citep{molit2021sidewalk}, the Seoul Sidewalk Design and Construction Manual, Version 3.0 \citep{seoul2026sidewalk}, and the National Survey of Pedestrian Transportation Conditions and Study on Its Utilization \citep{kotsa2022pedestrian}. The first two sources provide installation, design, and maintenance criteria, while the national survey treats effective sidewalk width and pavement condition as field-assessment items.

\begin{figure}[htbp]
    \centering
    \includegraphics[width=0.8\textwidth]{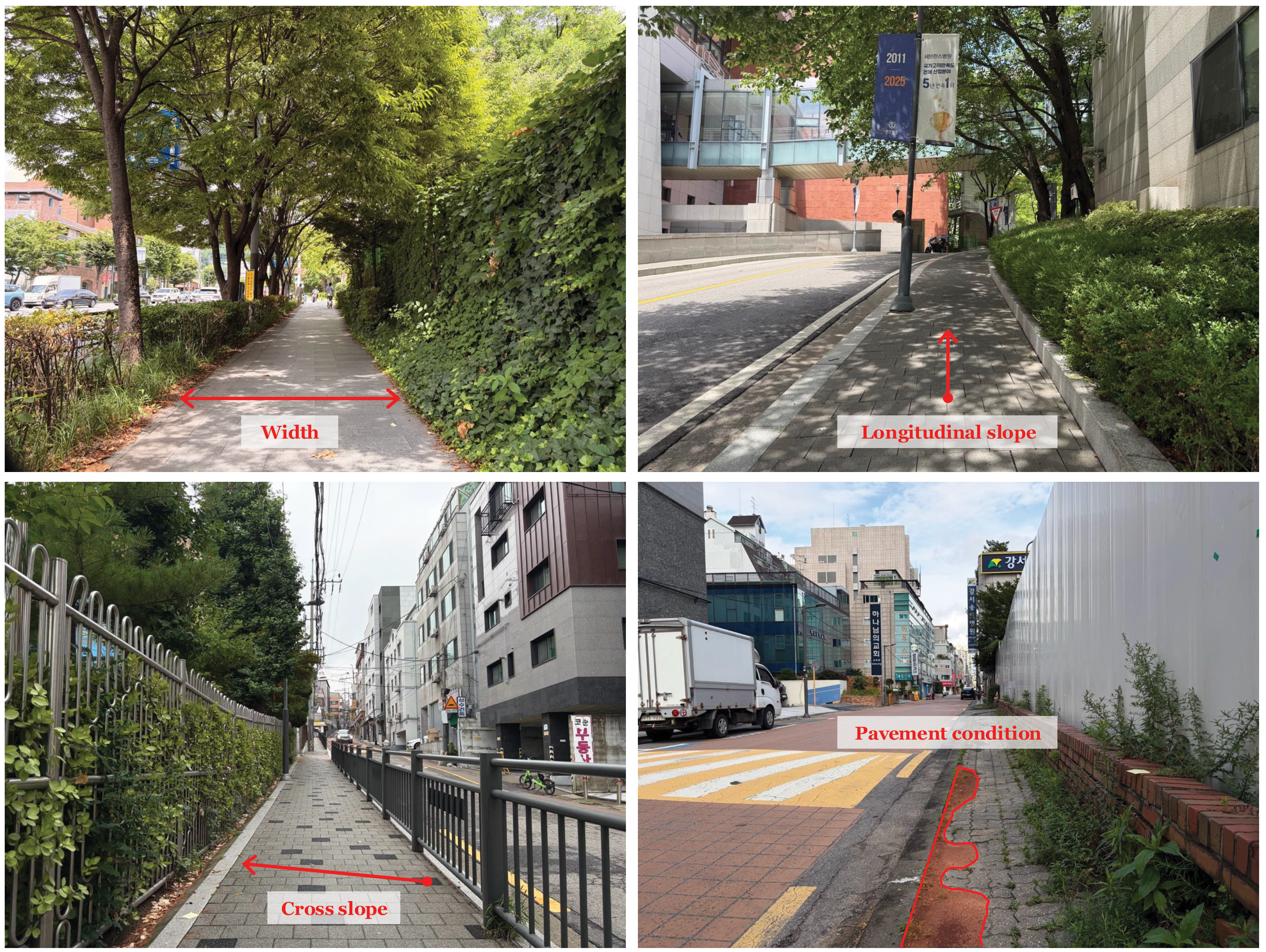}
    \caption{Representative field images of the four evaluated sidewalk accessibility attributes: effective width, longitudinal slope, cross slope, and pavement condition grade.}
    \label{fig:fig_2}
\end{figure}

The review identified six candidate sidewalk-accessibility attributes derived from Korean accessibility regulations: \textit{width, longitudinal slope, cross slope, pavement condition, curb ramp, and tactile paving.} Following the data collection (Section~\ref{subsec:datacollection}), curb ramps and tactile paving were excluded because almost all observed instances complied with the regulatory standards, leaving insufficient variability for a statistically meaningful evaluation. As a consequence, four attributes were retained for the CP study (Table \ref{tab:regulatory-thresholds}) with representative field examples shown in Figure \ref{fig:fig_2}. Collectively, these attributes capture complementary dimensions of sidewalk accessibility while covering both continuous (width and slopes) and discrete (pavement condition grade) prediction tasks, allowing evaluation of the proposed conformal framework across heterogeneous output spaces. 

The four retained attributes are regulation-derived, field-measurable components of sidewalk accessibility rather than a measure of accessibility as a whole. An overall accessibility evaluation would depend on factors beyond an image-based attribute assessment, including path continuity, temporal conditions, and network-level connectivity. Combining these dimensions into an overall accessibility judgment is outside the scope of this work. Our evaluation therefore focuses on each retained attribute independently.

\begin{table}[t]
\centering
\caption{Installation and maintenance criteria for the selected sidewalk accessibility attributes.}
\label{tab:regulatory-thresholds}
\begin{tabular}{p{0.22\linewidth} p{0.72\linewidth}}
\toprule
Attribute & Criterion \\
\midrule
Effective width &
At least 2.0 m. Where 2.0 m cannot be secured because of constraints associated with the expansion or improvement of an existing road, terrain, or physical obstructions, the width may be reduced to 1.5 m. Space occupied by street facilities and other obstructions is excluded from the effective width. \\

Longitudinal slope &
The longitudinal slope is generally limited to 1:18 (5.6\%, approximately $3.18^\circ$) and may be relaxed to 1:12 (8.3\%, approximately $4.76^\circ$) where terrain conditions make the general standard difficult to meet. \\

Cross slope &
The cross slope is generally limited to 1:50 (2.0\%, approximately $1.15^\circ$) and may be relaxed to 1:25 (4.0\%, approximately $2.29^\circ$) where the general standard cannot be met because of unavoidable terrain or surrounding conditions. \\

Pavement condition &
Pavement condition is classified from Grade A to Grade E according to its service condition. Sidewalks should be maintained at Grade C or better. \\
\bottomrule
\end{tabular}
\vspace{2pt}
\begin{minipage}{0.96\linewidth}
\footnotesize
\textit{Source:} Effective-width and slope criteria are drawn from \citep{molit2021sidewalk,seoul2026sidewalk}; the A--E pavement condition grades are drawn from \citep{molit2021sidewalk}. Slope ratios were converted to degrees using the arctangent of rise over run.
\end{minipage}
\end{table}

\begin{table}[b]
\centering
\caption{Overview of the data collection sites, including their administrative
district, collection period, and the number of images collected at each location.}
\label{tab:collection_sites}
\begin{tabular}{llll}
\toprule
\textbf{Site} & \textbf{District (Gu)} & \textbf{Collection period} & \textbf{No. of images} \\
\midrule
Seoul Senior Tower & Eunpyeong-gu & 8th July 2026 & 99 \\
Yonsei Severance Hospital & Seodaemun-gu & 7th July 2026 & 197 \\
Sinchon Montessori Kindergarten & Seodaemun-gu & 10th July 2026 & 100 \\
Siloam Center for the Blind & Mapo-gu & 11th July 2026 & 118 \\
\bottomrule
\end{tabular}
\end{table}

\subsection{Ground Truth Dataset Construction}\label{subsec:datacollection}
Following the taxonomy defined in Section \ref{subsec:taxonomi}, data collection combined standardized pedestrian-level image acquisition with manual field measurements, producing paired image–ground truth observations for each accessibility attribute retained in the study.

\subsubsection{Site Selection}
To capture a broad range of sidewalk accessibility conditions, data collection sites were selected across urban environments associated with different accessibility contexts. Site selection was informed by Article 2 of the Act on Promotion of the Transportation Convenience of Mobility Disadvantaged Persons \citep{korea2025mobility}, which identifies groups whose mobility needs should be considered in the provision of accessible transportation infrastructure. These groups informed the sampling strategy but were not themselves the object of assessment, as accessibility was evaluated as a property of the sidewalk environment. Accordingly, data were collected near facilities primarily serving older adults, persons with disabilities, persons accompanying infants, and persons with visual impairments.
Table \ref{tab:collection_sites} summarizes the selected sites together with their administrative district, collection period, and number of acquired images.

\subsubsection{Data Acquisition and Annotation}
Data collection followed a standardized field protocol to ensure consistency across all study locations. Images were acquired from the pedestrian perspective using an iPhone 17 (26 mm equivalent) and an iPhone 16 Pro (24 mm equivalent), positioned 1.0~m above ground level and aligned with the center of the pedestrian walking path. After image acquisition, each sidewalk segment was manually surveyed using standard field instruments to obtain the corresponding ground truth measurements, in line with prior sidewalk accessibility assessment methodologies \citep{ai2015automated}:

\begin{itemize}

\item \textit{Sidewalk width} was measured as the unobstructed pedestrian passage using a Sincon SD-70 laser distance meter, excluding permanent street furniture and other fixed obstacles according to the regulatory definition of effective width.

\item \textit{Longitudinal and cross slopes} were measured directly on the sidewalk surface using the same field instrumentation following the walking direction and the transverse direction, respectively. Cross slope was measured from the right edge to the left edge of the walking surface, matching the sign convention used in the model prompts (Appendix~\ref{app:prompts}).

\item \textit{Pavement condition grade} was annotated by a single member of the research team following the five-level (A–E) service-condition rubric defined in the Guidelines for Installation and Management of Sidewalks \citep{molit2021sidewalk}. Each image was assigned a single pavement condition grade, ranging from A (Very good) to E (Very poor), describing exclusively the condition of the walking surface. Grades were assigned based on visible indicators of pavement deterioration, including cracking, plastic deformation, settlement, uplift, potholes, and standing water, consistent with the five-level pavement condition grade scale (see Table~\ref{tab:pavement-grading}). 
\end{itemize}

\begin{table}[t]
\centering
\caption{Pavement condition grading rubric adopted for ground truth annotation.}
\label{tab:pavement-grading}
\begin{tabular}{p{0.08\linewidth} p{0.20\linewidth} p{0.62\linewidth}}
\toprule
\textbf{Grade} & \textbf{Condition} & \textbf{Description} \\
\midrule
A & Very good &
No pavement deformation or cracking; newly constructed or in like-new condition. \\
B & Good &
Minor surface irregularities may be present, but the pavement remains generally uniform and fully functional. \\
C & Fair &
Pedestrian passage remains possible, although resurfacing or maintenance may be required depending on the extent of deterioration. \\
D & Poor &
Pavement deterioration affects normal pedestrian passage; hazards are present on at least 50\% of the walking surface. \\
E & Very poor &
Normal pedestrian passage is no longer possible; hazards are present on at least 75\% of the walking surface. \\
\bottomrule
\end{tabular}
\end{table}

\begin{figure*}[htbp]
    \centering
\includegraphics[width=1.\textwidth]{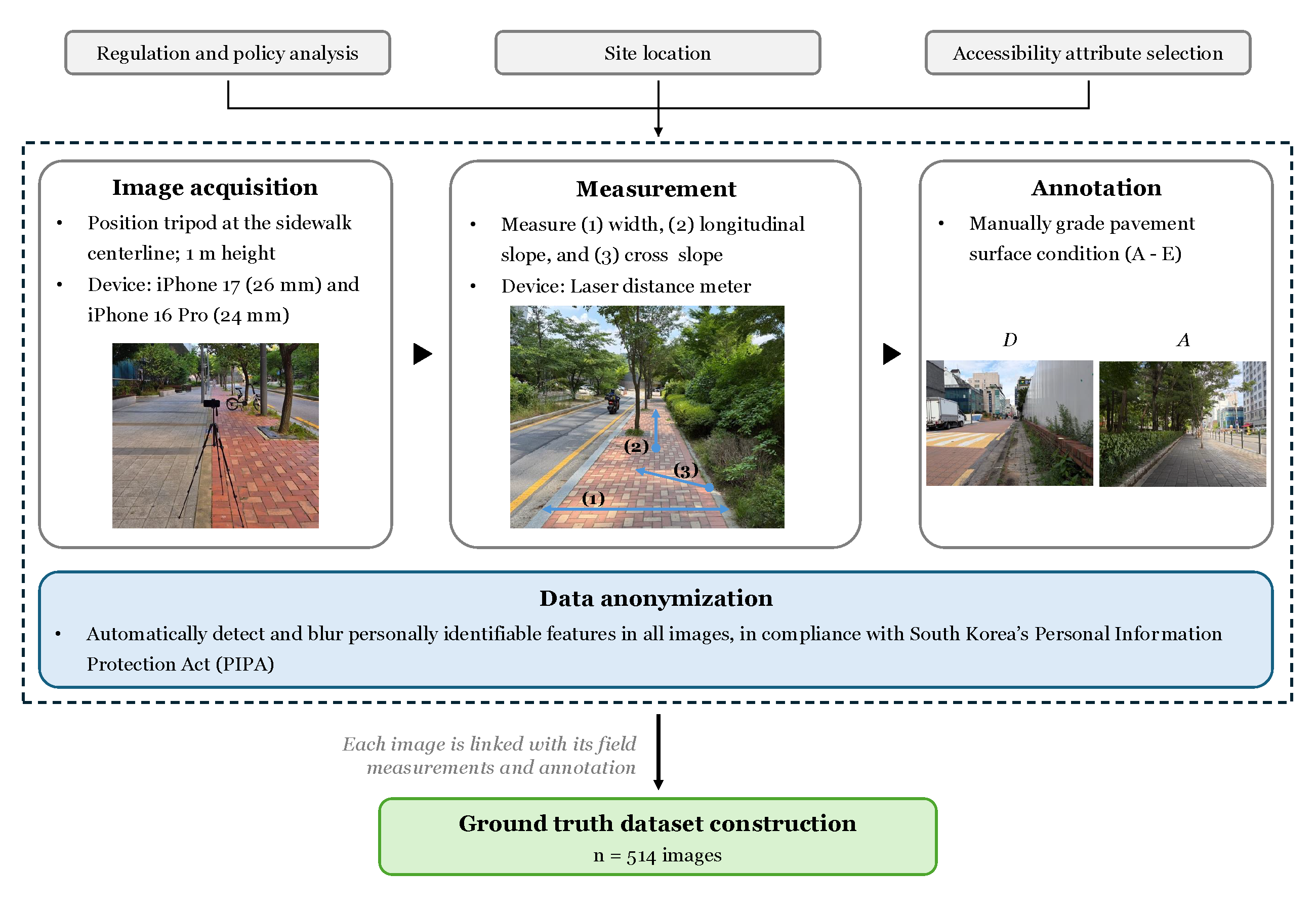}    \caption{Data collection process diagram.
}
    \label{fig:fig_3}
\end{figure*}

The resulting dataset comprises 514 annotated pedestrian-level sidewalk images collected during approximately 11 hours of fieldwork across the four study locations. Figure \ref{fig:fig_3} summarizes the complete data collection workflow.

\subsubsection{Data Anonymization}
All pedestrian-level imagery was collected and processed in accordance with South Korea’s Personal Information Protection Act (PIPA)\footnote{\href{https://elaw.klri.re.kr/eng_service/lawView.do?hseq=53044&lang=ENG}{Personal Information Protection Act (PIPA)}}, following the scientific research exemption and the data minimization principles established in Articles 2 and 3. Prior to model evaluation, all images underwent an anonymization procedure in which personally identifiable information was automatically detected and blurred. Only the anonymized images were retained for subsequent analysis. The resulting anonymized dataset does not intentionally retain personally identifiable information and preserves only the environmental context required for the study.

\subsection{Model Selection and Querying Protocol}
\label{sec:model-selection}

The conformal procedure of Section~\ref{sec:cp-framework} is model-agnostic. The choice of VLM affects the width of the calibrated regions rather than their validity, and it does so through a single channel. Both Equation~(\ref{eq:base}) and Equation~(\ref{eq:freq}) are functions of the dispersion of the $M$ sampled responses. A model whose repeated responses never vary consequently supplies no nonconformity signal. We therefore ran a pilot to fix the model set and to verify this precondition.

\paragraph{Candidates and versions}
We considered five model families, spanning proprietary API services and open-weight alternatives. Three are accessed through proprietary APIs (OpenAI GPT, Google Gemini, and Anthropic Claude). These families have consistently ranked at the top of large-scale human-preference evaluations of frontier models \citep{chiang2024chatbotarena}, so they are the systems practitioners are most familiar with and most likely to adopt, and none of them exposes logits. The remaining two are open-weight models at the locally deployable 8B scale, Qwen3-VL \citep{bai2025qwen3vl} and InternVL3.5 \citep{wang2025internvl35}, drawn from the open-weight suites that have been progressively closing the gap to commercial multimodal performance \citep{chen2024howfar}. Because these endpoints are versioned rapidly, we selected one version per family at the start of the pilot and used it unchanged throughout (Table~\ref{tab:models}). The findings reported below are properties of these versions.

\begin{table}[ht]
\centering
\caption{Evaluated vision-language models, held fixed for the duration of the study.}
\label{tab:models}
\small
\begin{tabular}{llll}
\toprule
Family & Version & Identifier & Access \\
\midrule
GPT          & GPT-5.2         & \texttt{gpt-5.2} & Proprietary API \\
Gemini       & Gemini 3 Flash  & \texttt{gemini-3-flash-preview}                      & Proprietary API \\
Claude       & Claude Opus 4.6 & \texttt{claude-opus-4-6}                    & Proprietary API, pilot only \\
Qwen3-VL     & 8B Instruct     & \texttt{Qwen/Qwen3-VL-8B-Instruct}                   & Open weights, local \\
InternVL3.5  & 8B              & \texttt{OpenGVLab/InternVL3\_5-8B-HF}                & Open weights, local \\
\bottomrule
\end{tabular}
\end{table}

\begin{figure}[t]
    \centering
    \includegraphics[width=\linewidth]{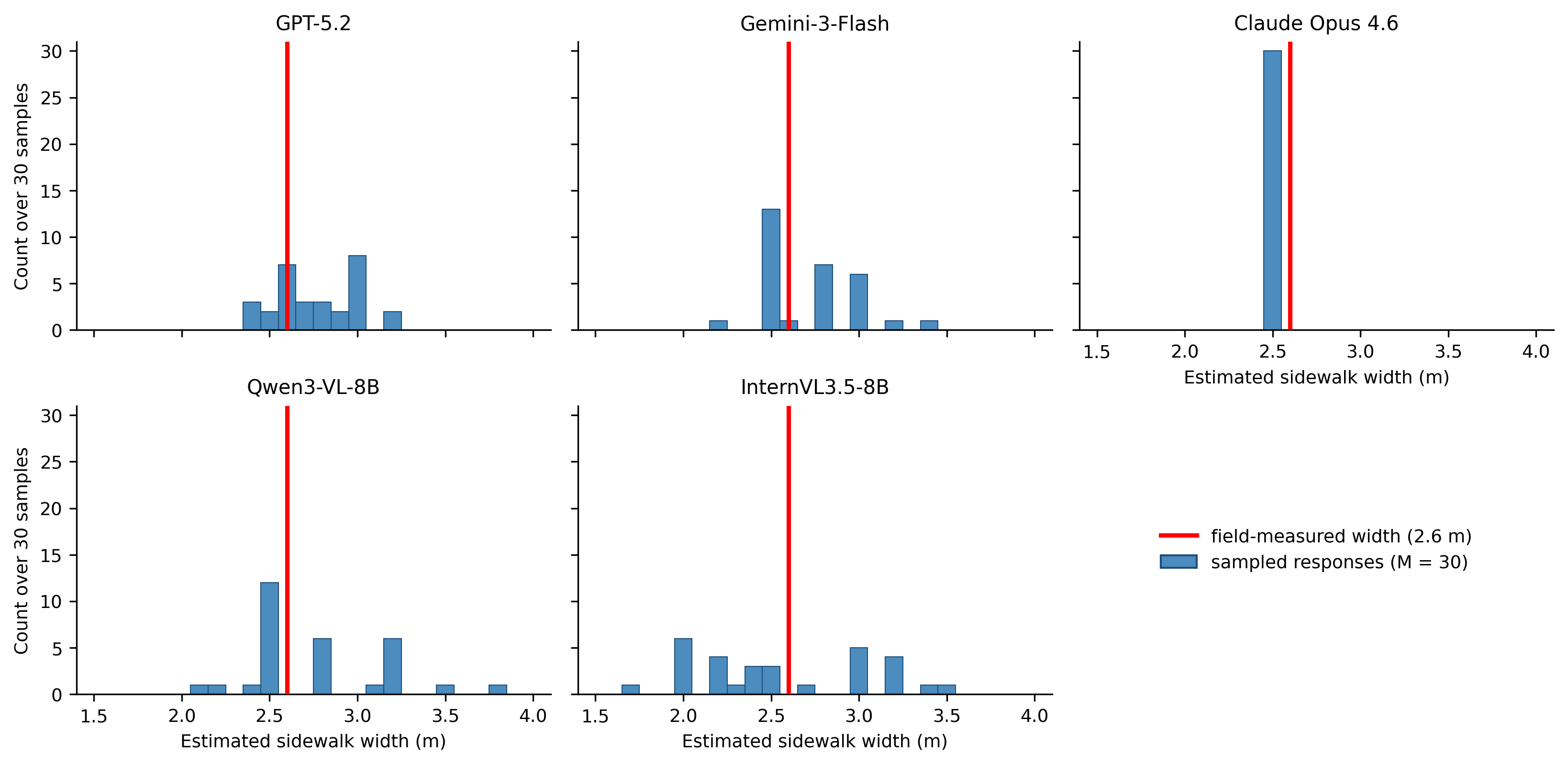}
    \caption{Empirical response distributions from 30 repeated sidewalk width queries on a single pilot image, shown per candidate model.}
    \label{fig:fig_4}
\end{figure}

\paragraph{Pilot and the exclusion of Claude} 
Ten images drawn at random from the 514 were queried $M=30$ times by each candidate on sidewalk width, under the configuration later used in the full study. Every request set temperature to $1.0$ with the remaining decoding parameters at their API defaults, used the same prompt and image encoding across candidates, and issued the $30$ queries as independent calls rather than a batched conversation. Claude Opus 4.6 returned an identical value across all 30 samples on every image (Figure~\ref{fig:fig_4}). While Claude's estimates lie as close to the field measurements as those of the other candidates, accuracy is not the criterion here. The pilot tests instead whether a model supplies the sampling variation the framework requires, and an invariant response cannot. The base interval collapses to a point, and the sampled frequencies become binary, so neither attribute type yields a usable nonconformity signal. Such invariance indicates deterministic decoding at the serving layer for this version rather than a property of the family. Therefore, the remaining four models proceeded to the full evaluation.

During the pilot we additionally tested two prompt formulations for sidewalk width, a concise instruction (Appendix~\ref{app:prompts}) and a longer variant with explicit measurement definitions and estimation rules (Appendix~\ref{app:prompt-alt}). Under repeated sampling, each model retained its characteristic distribution shape across the two formulations, but the response values shifted in location. Prompt wording therefore acts as a fixed component of the measurement instrument rather than a neutral interface. We fixed the concise formulation for all runs, and every calibrated result in this paper is conditional on it.

\paragraph{Querying protocol}
With Claude excluded, each request carried one image and one attribute prompt, with no in-context examples and no conversation history. The 30 responses for a given image are independent draws from the model's response distribution. The four attributes were queried in separate runs, and the response for one attribute did not condition the response for another. Decoding used temperature $1.0$ for all four models, with the API models at their remaining provider defaults and the open-weight models under nucleus sampling with top-p $0.95$. The full evaluation comprised 246{,}720 queries in total (4 models, 4 attributes, 514 images, 30 samples each). The two open-weight models ran locally on a single NVIDIA RTX A6000 (48\,GB). On a representative image, completing the 30 width queries took approximately 49.7\,s for GPT-5.2, 209.3\,s for Gemini-3-Flash, 25.9\,s for InternVL3.5-8B, and 20.3\,s for Qwen3-VL-8B, and the 30 pavement condition grading queries took 59.6, 162.9, 25.8, and 20.7\,s, respectively. The prompts issued for the four attributes are reproduced verbatim in Appendix~\ref{app:prompts}.

\section{Results}
\label{sec:results}

All results use the $n = 514$ annotated images, partitioned uniformly at random into a calibration half ($n_{\mathrm{cal}} = 257$) and a test half ($n_{\mathrm{test}} = 257$), with every reported quantity averaged over 200 independent partitions so that it reflects the calibration procedure rather than one fortunate split. The uniform split matches the exchangeability assumption under which the guarantee of Equation~(\ref{eq:coverage}) holds, so the reported coverage is valid for the surveyed population, though it does not test transfer to unseen locations, where spatial autocorrelation between nearby images breaks exchangeability. An area-stratified evaluation of that deployment setting is left to future work (Section~\ref{sec:limitations}).

Section~\ref{sec:res-data} first summarizes the field measurements that serve as ground truth. Sections~\ref{sec:res-width}--\ref{sec:res-cross} report the three continuous attributes under the rubric of Section~\ref{sec:eval-criteria}.  Section~\ref{sec:res-surface} reports the discrete pavement condition grade under the corresponding criteria.

\subsection{Field measurements}
\label{sec:res-data}

Table~\ref{tab:groundtruth} summarizes the field measurements over the 514 images. The surveyed sidewalks average 2.97~m in width (s.d.\ 1.36~m), yet the distribution is wide enough that 34.4\% of images fall below the 2.0~m standard of Table~\ref{tab:regulatory-thresholds} and 13.2\% fall below the 1.5~m constrained minimum. Both slopes are centered near level, with means of $-0.06^\circ$ (longitudinal) and $+0.29^\circ$ (cross) and standard deviations of 2.60$^\circ$ and 2.42$^\circ$, but the tails matter operationally, because segments with steep grades or pronounced surface tilt are precisely the cases an accessibility audit must identify. The pavement condition grades are heavily concentrated on intact pavement, with 80.2\% of images at grade A, no image at grade E, and only 3.9\% at grade C or worse.

\begin{table}[ht]
    \centering
    \caption{Field-measured ground truth over the 514 images.}
    \label{tab:groundtruth}
    \small
    \begin{tabular}{lcccc}
    \toprule
    Attribute & Mean & Std.\ dev. & Min. & Max. \\
    \midrule
    Effective width (m)           & 2.97    & 1.36 & 0.9    & 10.2 \\
    Longitudinal slope ($^\circ$) & $-0.06$ & 2.60 & $-9.3$ & 11.8 \\
    Cross slope ($^\circ$)        & 0.29    & 2.42 & $-9.7$ & 8.0  \\
    \midrule
    Pavement condition grade & \multicolumn{4}{l}{A\ 412 (80.2\%) \quad B\ 82 (16.0\%) \quad C\ 14 (2.7\%)
    \quad D\ 6 (1.2\%) \quad E\ 0} \\
    \bottomrule
\end{tabular}
\end{table}

\subsection{Width}
\label{sec:res-width}

\paragraph{Base intervals}
Before calibration, the base intervals cover the field-measured width on between 36.8\% and 45.5\% of images against a nominal 90\% (Table~\ref{tab:width}). The dispersion of repeated responses therefore understates the true error by a wide margin, extending to the sampled-quantile route the overconfidence that \citet{epstein2025llms} document for directly elicited confidence intervals.

\paragraph{Validity}
After calibration, empirical coverage lies between 0.905 and 0.915 across all models and both variants, and tracks the nominal level over the full sweep $\alpha \in \{0.05, \dots, 0.30\}$. Coverage marginally above the $1 - \alpha + \tfrac{1}{n_{\mathrm{cal}}+1} = 0.904$ upper bound of \citet{romano2019quantile} is expected here, because responses on a 0.1~m grid produce heavily tied nonconformity scores and the distinctness condition of the bound fails. Since validity is assured for every model, the models are compared only on interval width.

\paragraph{Bias structure and calibration variants}
The measured width falls below the base interval on 50.0\% to 59.9\% of images but above it on only 1.8\% to 7.6\%, giving ratios $\pi_{lo}/\pi_{hi}$ between 6.6 and 29.3. Correspondingly, the median sampled response exceeds the measured width by $+0.40$~m for GPT-5.2 up to $+1.05$~m for InternVL3.5-8B, so all four models systematically overestimate how wide a sidewalk is. Because the misses concentrate in one tail, the pooled correction of Equation~(\ref{eq:symmetric}) displaces the upper endpoint further than coverage requires, and the symmetric variant spends its risk budget almost entirely on the lower tail. After symmetric calibration, misses split into 0.090 below against 0.004 above the interval for Gemini-3-Flash. Calibrating each endpoint separately restores an even split at $\alpha/2$ per side and shortens mean intervals by 5.7\% (GPT-5.2), 20.6\% (Gemini-3-Flash), 25.7\% (Qwen3-VL-8B) and 33.2\% (InternVL3.5-8B) at unchanged coverage, and the gain grows with the size of the bias.

For example, on a test image where GPT-5.2 (whose asymmetric intervals are the narrowest of the four models) most frequently answers 2.0 m, its 30 responses give a band of [0.75 m, 2.77 m] at the calibrated 90\%. The band is anchored on the base interval of the 30 sampled responses, whose endpoints are extended by the per-tail corrections $\hat q_{lo} = 0.80$~m below and $\hat q_{hi} = 0.33$~m above, so the larger allowance sits under the lower endpoint, exactly where the overestimation bias places the misses. An assessment workflow receives this band, not the 2.0 m point value.

\paragraph{Self-consistent errors}
The diagnostic restricts attention to the images where the model is most confident. Within the narrowest quartile of base intervals, whose mean interval width is under 0.5~m, the field measurement escapes the base interval on 77\% to 90\% of images against a nominal 10\%. Agreement across repeated samples is therefore no evidence of correctness, which is the continuous form of the self-consistent-error regime documented by \citet{tan2025consistent}, and coverage on these confidently wrong images is achieved only through the uniform conformal correction.

\paragraph{Decision relevance}
The asymmetric intervals correspond to mean half-widths of roughly $\pm 1.0$~m (GPT-5.2) to $\pm 1.5$~m (InternVL3.5-8B). Read against the 2.0~m standard and the 1.5~m constrained minimum of Table~\ref{tab:regulatory-thresholds}, a segment can be classified as compliant or non-compliant only when its entire interval lies on one side of the threshold.
Classification is therefore confined to segments lying sufficiently far from the regulatory thresholds, while intervals for observations closer to either threshold remain inconclusive. Although width is the most informative quantitative attribute evaluated, the approximately $\pm 1.0$~m uncertainty of the best-performing model is too large for general-purpose compliance assessment and limits automated classification to extreme cases.

\begin{table}[ht]
\centering
\caption{Effective width at $\alpha = 0.10$, averaged over 200 calibration--test partitions
(257/257).}
\label{tab:width}
\small
\begin{tabular}{lcccccccc}
\toprule
& \multicolumn{2}{c}{Base interval} & \multicolumn{2}{c}{Violations}
& \multicolumn{2}{c}{Symmetric} & \multicolumn{2}{c}{Asymmetric} \\
\cmidrule(lr){2-3}\cmidrule(lr){4-5}\cmidrule(lr){6-7}\cmidrule(lr){8-9}
Model & cov. & width (m) & $\pi_{lo}$ & $\pi_{hi}$ & cov. & width (m) & cov. & width (m) \\
\midrule
GPT-5.2        & 0.424 & 0.89 & 0.500 & 0.076 & 0.907 & 2.13 & 0.910 & 2.01 \\
Gemini-3-Flash & 0.368 & 1.28 & 0.599 & 0.033 & 0.906 & 2.87 & 0.910 & 2.27 \\
Qwen3-VL-8B    & 0.393 & 1.32 & 0.562 & 0.045 & 0.907 & 3.71 & 0.915 & 2.75 \\
InternVL3.5-8B & 0.455 & 2.47 & 0.527 & 0.018 & 0.905 & 4.59 & 0.909 & 3.07 \\
\bottomrule
\end{tabular}
\end{table}

\subsection{Longitudinal slope}
\label{sec:res-running}

\paragraph{Base intervals}
Base coverage falls to between 32.9\% and 47.1\% (Table~\ref{tab:running}), with base widths of 1.6$^\circ$ to 2.1$^\circ$ against sample-median absolute errors of 1.5$^\circ$ to 1.9$^\circ$. The sampled dispersion is again far too tight to describe the true error.

\paragraph{Validity}
Conformalization restores coverage of 0.902 to 0.908 for every model and variant, tracking the nominal level across the sweep, at the cost of intervals of 7.1$^\circ$ to 9.4$^\circ$. As for width, tied scores keep empirical coverage at or marginally above the nominal level, and the models are separated only by interval width.

\paragraph{Bias structure and calibration variants}
Unlike width, the longitudinal slope is a signed quantity distributed about zero, and the violation ratios $\pi_{lo}/\pi_{hi}$ collapse to between 0.8 and 2.4. With no tail to exploit, asymmetric calibration changes mean width by between $-1.2$\% and $+4.0$\%, and for three of the four models it lengthens the intervals, exactly the behavior \citet{romano2019quantile} report when the per-tail guarantee is bought without a compensating bias. This confirms, on field data, the prediction of Section~\ref{sec:cp-framework} that the direction of the symmetric--asymmetric comparison is an empirical property of the attribute rather than of the method.

\paragraph{Self-consistent errors}
The diagnostic repeats the width pattern. Within the narrowest quartile of base intervals, the measurement escapes on 53\% to 85\% of images against a nominal 10\%.

\paragraph{Decision relevance}
The calibrated intervals span 7.11$^\circ$ (GPT-5.2) to 9.04$^\circ$ (InternVL3.5-8B), corresponding to half-widths of about $\pm 3.6^\circ$ to $\pm 4.5^\circ$. These half-widths are of the same order as the regulatory limits themselves (3.18$^\circ$ general, 4.76$^\circ$ relaxed; Table~\ref{tab:regulatory-thresholds}). An interval centered near level spans both limits, and only pronouncedly steep segments can be placed on one side of them. Longitudinal slope is marginal for compliance screening at this confidence level.

\begin{table}[ht]
\centering
\caption{Longitudinal slope at $\alpha = 0.10$, same protocol as Table~\ref{tab:width}.}
\label{tab:running}
\small
\begin{tabular}{lcccccccc}
\toprule
& \multicolumn{2}{c}{Base interval} & \multicolumn{2}{c}{Violations}
& \multicolumn{2}{c}{Symmetric} & \multicolumn{2}{c}{Asymmetric} \\
\cmidrule(lr){2-3}\cmidrule(lr){4-5}\cmidrule(lr){6-7}\cmidrule(lr){8-9}
Model & cov. & slope ($^\circ$) & $\pi_{lo}$ & $\pi_{hi}$ & cov. & slope ($^\circ$) & cov. & slope ($^\circ$) \\
\midrule
GPT-5.2        & 0.329 & 1.59 & 0.471 & 0.200 & 0.903 & 7.11 & 0.907 & 7.14 \\
Gemini-3-Flash & 0.360 & 1.66 & 0.389 & 0.251 & 0.904 & 7.32 & 0.908 & 7.23 \\
Qwen3-VL-8B    & 0.354 & 2.04 & 0.290 & 0.356 & 0.903 & 8.28 & 0.907 & 8.51 \\
InternVL3.5-8B & 0.471 & 2.12 & 0.265 & 0.265 & 0.902 & 9.04 & 0.905 & 9.39 \\
\bottomrule
\end{tabular}
\end{table}

\subsection{Cross slope}
\label{sec:res-cross}

\paragraph{Base intervals}
Base coverage spans 16.9\% to 39.9\% (Table~\ref{tab:cross}). The extreme is Gemini-3-Flash, whose sampled responses concentrate so tightly around zero (mean base width $0.71^\circ$) that the base interval misses the measurement on 83\% of images. Cross slope is where base overconfidence reaches its worst value in the study.

\paragraph{Validity}
Conformalization restores coverage of 0.905 to 0.912 for every model and variant, at the cost of intervals of 8.2$^\circ$ to 10.0$^\circ$, and the sweep tracks the nominal level as before.

\paragraph{Bias structure and calibration variants}
The violations now lean the other way, with $\pi_{hi} > \pi_{lo}$ for every model, reflecting a tendency to predict below the measured value, but the imbalance is mild (ratios 0.4 to 0.6) and asymmetric calibration accordingly moves mean width by only $-1.8$\% to $+1.7$\%. Together with Section~\ref{sec:res-running} this completes the pattern anticipated in Section~\ref{sec:cp-framework}. The asymmetric gain is large exactly where the bias is one-sided and disappears where it is not.

\paragraph{Self-consistent errors}
The confidently-wrong diagnostic is at its most extreme here, with GPT-5.2 missing on 96\% of its narrowest-quartile images.

\paragraph{Decision relevance}
The calibrated intervals span 8.21$^\circ$ (Gemini-3-Flash) to 9.97$^\circ$ (GPT-5.2), corresponding to half-widths of $\pm 4.1^\circ$ to $\pm 5.0^\circ$, against regulatory limits of 1.15$^\circ$ (general) and 2.29$^\circ$ (relaxed). Every interval therefore spans the entire regulatory range several times over. No images can be certified compliant or non-compliant at this confidence level, and cross slope is not deployable for screening. Coverage is nonetheless maintained, so validity is preserved; the limitation appears as interval widths too large to resolve the thresholds.

\begin{table}[ht]
\centering
\caption{Cross slope at $\alpha = 0.10$, same protocol as Table~\ref{tab:width}.}
\label{tab:cross}
\small
\begin{tabular}{lcccccccc}
\toprule
& \multicolumn{2}{c}{Base interval} & \multicolumn{2}{c}{Violations}
& \multicolumn{2}{c}{Symmetric} & \multicolumn{2}{c}{Asymmetric} \\
\cmidrule(lr){2-3}\cmidrule(lr){4-5}\cmidrule(lr){6-7}\cmidrule(lr){8-9}
Model & cov. & slope ($^\circ$) & $\pi_{lo}$ & $\pi_{hi}$ & cov. & slope ($^\circ$) & cov. & slope ($^\circ$) \\
\midrule
GPT-5.2        & 0.395 & 2.16 & 0.175 & 0.430 & 0.906 & 9.97 & 0.907 & 9.79 \\
Gemini-3-Flash & 0.169 & 0.71 & 0.321 & 0.510 & 0.908 & 8.21 & 0.912 & 8.27 \\
Qwen3-VL-8B    & 0.379 & 1.78 & 0.212 & 0.409 & 0.906 & 8.32 & 0.910 & 8.34 \\
InternVL3.5-8B & 0.399 & 1.94 & 0.216 & 0.385 & 0.905 & 8.49 & 0.909 & 8.64 \\
\bottomrule
\end{tabular}
\end{table}

\subsection{Pavement condition grade}
\label{sec:res-surface}

\paragraph{Degeneracy}
The discrete attribute realizes the degeneracy regime of Section~\ref{sec:cp-framework} rather than simply risking it. The field distribution is heavily skewed toward intact pavement (412 of 514 images at grade A, none at E), and three of the four models almost never emit grade A, shifting pristine segments one notch to B. GPT-5.2 assigns modal grade B to 383 of the 412 grade-A images, Qwen3-VL-8B to 341 and InternVL3.5-8B to 305. Their sampling distributions are also nearly deterministic, with mean modal mass 0.72 to 0.97 concentrated on 1.2 to 2.0 of the five grades, so the errors are self-consistent. The proportion of calibration images whose true grade is never sampled is consequently $\rho = 0.70$ for GPT-5.2, $0.83$ for Qwen3-VL-8B and $0.81$ for InternVL3.5-8B. Since $\rho > \alpha$ at every level tested, Equation~(\ref{eq:degeneracy}) forces $\hat q = 1$ on every one of the 200 partitions. The prediction set is then all five grades for every image, coverage is trivially 100\%, and top-1 accuracy of 10.1\% to 20.6\% confirms there is no signal for the threshold to find (Table~\ref{tab:grade}).

\paragraph{Behavior near the boundary}
Gemini-3-Flash, with $\rho = 0.095$ and top-1 accuracy 62.3\%, sits almost exactly on the $\rho = \alpha$ boundary at the working level, and its behavior there is a finding about methodology as much as about the model. At $\alpha = 0.10$, 52\% of individual partitions are degenerate and 48\% are not. The split-averaged summary (coverage 0.950, mean set size 3.50) thus describes a mixture that occurs on no actual partition, the two components being full-set behavior at size 5.0 and informative behavior near 1.8 grades. Averages over repeated splits, standard in conformal evaluation, are therefore misleading whenever $\rho$ sits near $\alpha$, and the degenerate-split fraction should be reported alongside them. One step down the confidence ladder, the mixture resolves. At $\alpha = 0.15$ no partition is degenerate and the sets average 1.76 grades at coverage 0.879, at $\alpha = 0.20$ they average 1.51 grades at 0.812, and at $\alpha = 0.25$ they reach 1.31 grades at 0.753. Sharper sets are therefore available for this model, but each step is paid for in guaranteed coverage.

\begin{table}[ht]
\centering
\caption{Pavement condition grade across confidence levels, averaged over 200 calibration--test partitions (257/257). $\rho$ is the proportion of calibration images whose true grade is never sampled and top-1 the modal-grade accuracy, both independent of $\alpha$. Deg.\ frac.\ is the fraction of partitions with $\hat q = 1$.}
\label{tab:grade}
\small
\begin{tabular}{lccccccc}
\toprule
Model & $\rho$ & top-1 & $\alpha$ & $\hat q$ (mean) & deg.\ frac. & cov. & set size \\
\midrule
GPT-5.2        & 0.696 & 0.206 & 0.10--0.25 & 1.000 & 1.00 & 1.000 & 5.00 \\
\midrule
Gemini-3-Flash & 0.095 & 0.623 & 0.10 & 0.984 & 0.52 & 0.950 & 3.50 \\
               &       &       & 0.15 & 0.950 & 0.00 & 0.879 & 1.76 \\
               &       &       & 0.20 & 0.876 & 0.00 & 0.812 & 1.51 \\
               &       &       & 0.25 & 0.774 & 0.00 & 0.753 & 1.31 \\
\midrule
Qwen3-VL-8B    & 0.831 & 0.109 & 0.10--0.25 & 1.000 & 1.00 & 1.000 & 5.00 \\
InternVL3.5-8B & 0.807 & 0.101 & 0.10--0.25 & 1.000 & 1.00 & 1.000 & 5.00 \\
\bottomrule
\end{tabular}
\end{table}

\paragraph{Screening against the maintenance threshold}
Read against the regulatory requirement that pavement be maintained at grade C or better, the degenerate models are inert. A five-grade set straddles the boundary between acceptable (A to C) and maintenance-needed (D and E) on every image, so GPT-5.2, Qwen3-VL-8B and InternVL3.5-8B settle 0\% of screening decisions at every level tested and every image would be routed to inspection. Gemini-3-Flash is the sole model that decides anything. Its decisive fraction rises from 47\% at $\alpha = 0.10$ to essentially 100\% at $\alpha \ge 0.15$, with a false-clear rate near 1\%. This last figure must be read against the field distribution rather than as evidence of skill, since only 6 of the 514 images fall at grade D or worse and the fixed set of grades A to C alone covers 98.8\% of field grades. The low false-clear rate therefore mainly mirrors the scarcity of damaged pavement, and the data cannot establish that any model detects the segments that actually need maintenance. What CP does establish, from calibration data alone, is that three of the four models provide no usable information on this attribute, before any deployment would reveal it.

\subsection{Computational cost and query time}
\label{sec:cost}

We measured per-image cost on a representative image under the production configuration of Section~\ref{sec:study-design} for the width and pavement condition attributes. Sampling one image 30 times for the width attribute costs approximately \$0.03 for GPT-5.2 and \$0.12 for Gemini-3-Flash, or roughly \$15 and \$62 for the full 514-image width assessment. The two slope attributes use prompts of comparable length to width. The pavement condition task costs \$0.04 for GPT-5.2 and \$0.08 for Gemini-3-Flash. The open-weight models carry no per-query fee.

From this preliminary analysis, the price of the coverage guarantee is the sampling budget $M$ itself, and reducing $M$ lowers cost but coarsens the frequency-based scores. Moreover, the models that are free to query locally are also the ones that produce the widest intervals. The practical choice is therefore between paying per query for narrower regions and paying in hardware for wider ones.

\section{Discussion}\label{sec:discussion}

\subsection{Synthesis across attributes}
\label{sec:synthesis}
Our results show that the relevant question for VLM-based accessibility assessment, posed in Section~\ref{sec:intro}, is not whether a model is accurate on average. It is whether the calibrated interval or set of each attribute is narrow enough to determine on which side of the regulatory threshold a sidewalk lies. Coverage itself is supplied by the calibration procedure regardless of model quality. We therefore compare the calibrated intervals and sets with the regulatory thresholds of Table~\ref{tab:regulatory-thresholds}, attribute by attribute.

\paragraph{Width}

Sidewalk width provides the most informative quantitative intervals among the evaluated attributes. At 90\% confidence, GPT-5.2 bounds it to a mean half-width of about 1.0~m, smaller than the 1.36~m spread of the surveyed widths but still substantial relative to the 1.5~m and 2.0~m regulatory thresholds.
Threshold position can therefore be resolved only for segments lying sufficiently far from the relevant boundary, while observations closer to either threshold remain inconclusive.
While width is comparatively more informative than the other quantitative attributes, its uncertainty remains too large for general quantitative compliance assessment.
Width is also the attribute for which the choice of calibration variant matters most: every model overestimates width, so calibrating the two tails separately shortens the intervals by 5.7\% to 33.2\% at unchanged coverage. A plausible reading of the bias is that the models estimate the full paved span between curb and building line rather than the unobstructed pedestrian passage that the regulatory definition requires, which excludes street furniture and fixed obstacles. This reading is consistent with the one-sided violation pattern ($\pi_{lo} \gg \pi_{hi}$, Table~\ref{tab:width}) and explains why per-tail calibration is so effective here, as nearly the entire correction is spent on the lower endpoint while the upper correction remains small.

\paragraph{Slopes}
Neither slope is precise enough for threshold screening, though the two differ in degree.
Longitudinal slope is bounded to about $\pm 3.6^\circ$ at best, a half-width larger than the general regulatory limit (3.18$^\circ$), close to the relaxed limit (4.76$^\circ$), and larger than the 2.60$^\circ$ spread of the field distribution. An interval centered near level spans both limits and only pronounced grades escape it. The narrow base intervals suggest that responses cluster near level regardless of the visible gradient, plausibly because perspective foreshortening in forward-facing imagery makes gentle grades appear flat, and the self-consistent-error diagnostic points the same way, since the images on which a model is most certain are disproportionately those on which it is most wrong. Cross slope is bounded no tighter than $\pm 4.1^\circ$ against limits of 1.15$^\circ$ and 2.29$^\circ$ and a field spread of 2.42$^\circ$. Every interval spans the regulatory range several times over and no compliance determination is possible. The acquisition geometry compounds the difficulty, as the tilt perpendicular to the walking direction is barely expressed in imagery captured along the sidewalk centerline, and the extreme concentration of Gemini-3-Flash's responses near zero (mean base width $0.71^\circ$) indicates responses that are essentially uninformative about this dimension. In both cases coverage is maintained throughout, and the failure is expressed where it belongs, in widths too coarse for the thresholds that matter.

\paragraph{Pavement condition grade}
The pavement condition grade fails before calibration: the models make a consistent one-grade error rather than random errors. Three of the four models assign grade B to pristine pavement, plausibly reading the minor surface texture visible even in new pavement as a defect, and almost never sample the true grade. Consequently, their calibrated sets degenerate to all five grades at every level tested and settle no screening decision at all. Because the error is self-consistent, the sampling distribution alone cannot distinguish a confidently wrong grade from a confidently correct one.
Only comparison with the field grade during calibration reveals the error: images whose true grade is never sampled receive the maximal nonconformity score, and once these exceed a fraction $\alpha$ of the calibration set the threshold admits all five grades.
Gemini-3-Flash is the only exception, and only at $\alpha \ge 0.15$, where its sets shrink to 1.3 to 1.8 grades and decide nearly all maintenance cases with a false-clear rate near 1\%.

\paragraph{Model comparison}
GPT-5.2 produces the narrowest calibrated intervals for width (2.01~m) and longitudinal slope ($7.11^\circ$), but the widest cross-slope intervals (9.97$^\circ$) and fully degenerate grading. Gemini-3-Flash is the most consistent model across attributes, second on width and longitudinal slope, first on cross slope (8.21$^\circ$), and the only model whose grade sets do not degenerate ($\rho = 0.095$ against 0.70 to 0.83 for the others, with informative sets of 1.3 to 1.8 grades at $\alpha \ge 0.15$). The two open-weight 8B models never lead on any attribute, and InternVL3.5-8B produces the widest longitudinal-slope intervals (9.04$^\circ$). 
As every model attains nominal coverage, this is a ranking of interval width and set size at equal coverage. The proprietary and open-weight models differ in the width of their calibrated regions, not in validity.

\paragraph{Regularities across attributes}
Whether a continuous attribute can be screened at a given confidence level depends on two comparisons. The first compares the calibrated half-width with the regulatory limit it must resolve, and the second compares the calibrated half-width with the spread of the field distribution. Width is the only attribute whose half-width is smaller than the field spread, but it remains large relative to the 1.5~m and 2.0~m thresholds. Longitudinal-slope half-widths are of the same order as the regulatory limits and larger than the field spread, and cross-slope half-widths exceed both by a wide margin. Pavement condition grading fails for a different reason, since three of four models never sample the true grade. Beyond this attribute-by-attribute picture, two regularities hold across all four attributes. First, validity is universal. Every model, attribute, and calibration variant attains its nominal coverage, so coverage does not distinguish the models. Second, overconfidence before calibration is universal. Raw sampling dispersion covers 17\% to 47\% of measurements against a nominal 90\%, and the intervals miss most often where the samples agree most closely, so response consistency cannot be read as confidence. The practical consequence is that whether a VLM assessment is precise enough for a regulatory decision must be evaluated per attribute rather than per model. The same model resolves the threshold for width on segments far from it and fails to resolve the threshold for cross slope on every segment.

\subsection{Implications for Municipal Sidewalk Accessibility Auditing}\label{sec:implications}
The proposed framework should not be interpreted as replacing professional accessibility audits. Rather, it provides a statistically grounded way to evaluate whether VLM predictions are sufficiently precise for a specific screening decision. Under the conditions evaluated here, none of the quantitative attributes supports general regulatory compliance assessment at 90\% coverage.

This distinction between validity and precision sufficient for a decision is important for municipal use. Conformal calibration guarantees that the prediction regions attain the specified marginal coverage, but it does not guarantee that those regions are sufficiently narrow to support a regulatory decision. Calibration data can therefore be used before operational deployment to determine whether the uncertainty associated with a given attribute is compatible with the intended screening task. Attributes whose calibrated intervals remain too wide relative to the relevant regulatory thresholds, or whose prediction sets degenerate to the full label space, can be identified as insufficiently informative rather than being converted into unsupported automated decisions. In this sense, the immediate value of the framework is not to automate accessibility auditing, but to provide an empirical basis for determining which VLM-based assessments, if any, have reached the precision required for a particular decision.

The results also caution against using simpler uncertainty heuristics in place of calibration. A seemingly natural alternative would be to route reports according to agreement among a model's repeated responses, treating tightly clustered samples as trustworthy and dispersed responses as requiring review. Such a rule would require no field-measured calibration data and corresponds to the self-consistency heuristics reviewed in Section~\ref{sec:related-reliability}. However, the self-consistent-error diagnostic in Section~\ref{sec:results} shows that response agreement does not indicate correctness. Tightly clustered responses can be  systematically wrong, and the images with the most self-consistent responses miss the field-measured value at high rates. A consistency-based screening rule would therefore assign high confidence to exactly the predictions that are systematically biased. Calibration against field-measured ground truth is consequently a prerequisite for interpreting repeated VLM responses as uncertainty estimates in compliance-oriented applications.

The computational cost of this calibration-based evaluation is comparatively modest. Under the production configuration described in Section~\ref{sec:cost}, assessing a single attribute with the full sampling budget of $M = 30$ costs \$0.03--\$0.12 per image for the API models, while the open-weight models incur no per-query fee. A four-attribute assessment would therefore remain below one US dollar per image at the observed API prices. These figures characterize inference costs rather than deployment feasibility: the present results show that low computational cost does not compensate for prediction regions that remain too wide for the intended regulatory decision.

Field-based accessibility assessment, by contrast, requires substantially more human effort. Trained agency reviewers have been reported to require approximately two hours per mile for pathway data collection \citep{zhang2023oasis}, while collection of the field measurements used as ground truth in this study required approximately 11 hours. Pedestrian-level imagery could reduce acquisition costs when suitable imagery is already available, including in citizen-reporting workflows such as the motivating scenario described in Section~\ref{sec:intro}. However, our results do not establish that current VLM predictions can replace these field measurements for quantitative compliance assessment. Field-measured data remain necessary both to calibrate the prediction regions and to resolve cases for which those regions are insufficiently informative. The practical role of calibration is therefore to quantify this boundary explicitly, rather than to assume that scalable image acquisition translates directly into reliable automated assessment.

\subsection{Relation to Prior Work on Conformal Prediction for Black-Box VLMs}\label{sec:prior-cp}

Our results relate to the sampling-based CP literature reviewed in Section~\ref{sec:related-cp} in three ways. The first concerns validity, where our coverage results agree with those reported in prior work. \citet{su2024APIconformal} obtain nominal coverage from frequency-based scores on text question-answering benchmarks, both open-ended and multiple-choice, \citet{yang2025conformalsets} on multiple-choice question answering, and \citet{ye2025datadrivencalibration} on multimodal question answering. We reproduce nominal coverage in every configuration tested, spanning four models, four attributes, and, for the continuous attributes, both calibration variants, on continuous and ordinal outputs anchored to regulatory thresholds and validated against field measurements rather than benchmark labels. The coverage guarantee holds for field-measured urban data as the theory predicts.

The second point is where our results qualify the literature, namely informativeness. Prior evaluations, including the logit-based vision study of \citet{fillioux2026foundationmodels}, which concludes that foundation models conformalize well, operate on benchmarks where the underlying models are already reasonably accurate, so valid coverage arrives with usably small sets. In our setting, the same procedure spans a broad range of informativeness, from comparatively informative but still limited width estimates, through marginal longitudinal-slope estimates and valid but uninformative cross-slope intervals, to fully degenerate pavement condition grade sets for three of four models. The score convention from which degeneracy follows, a maximal score for calibration labels the model never samples, originates with \citet{su2024APIconformal}. Our results show that the full-set regime this convention implies is realized in practice, and Section~\ref{sec:discrete} characterizes when, namely whenever the unsampled fraction exceeds the risk level ($\rho > \alpha$). Behavior near the $\rho = \alpha$ boundary adds a further caution. Split-averaged summaries there describe a mixture realized on no single partition, and the degenerate-split fraction should be reported alongside them.

The third concerns sampling dispersion as an uncertainty signal in its own right, where our evidence is uniformly negative. The overconfidence that \citet{epstein2025llms} document for elicited confidence intervals reappears in sampled-quantile intervals, whose base coverage reaches only 17\% to 47\% against a nominal 90\%. The self-consistent errors of \citet{tan2025consistent} acquire a continuous, per-image instantiation.
The implication for dispersion-based scores, whether computed from response frequency or from the entropy of semantically clustered responses \citep{wang2023selfconsistency, zhang2024vluncertaintydetecting}, is direct, as no function of the samples can lower the score of a value the model never produces.
Only calibration against ground truth converts that blindness into wide or degenerate regions instead of confident error. In sum, sampling-based CP transfers to heterogeneous, regulation-anchored, field-validated outputs with its guarantee intact, and the open problems shift from validity to informativeness and to coverage under the spatial structure of deployment data, for which exchangeability-aware extensions \citep{wang2025sconu, lin2026domainshift} offer a starting point.

\subsection{Generalizability to Other Accessibility Attributes}\label{sec:generalizability}

The framework of Section~\ref{sec:cp-framework} places no constraint on the attribute type. Any attribute for which a nonconformity score can be computed from repeated samples and a calibration set with field measurements can be assembled is a valid candidate, and the results indicate what to expect from the two attribute families the present study did not evaluate.

Curb ramps and tactile paving, both defined in the same regulatory sources as the retained attributes \citep{molit2021sidewalk,seoul2026sidewalk}, were excluded because nearly all observed instances complied with the standards (Section~\ref{subsec:taxonomi}). For such binary presence attributes the frequency score of Section~\ref{sec:discrete} reduces to the fraction of the $M$ responses that report the feature as present, and the compliance skew that motivated the exclusion creates two distinct risks. If the model systematically mis-reads the prevalent class, the true label goes unsampled on more than a fraction $\alpha$ of calibration images and the sets degenerate to the full label space, the outcome observed for three models on pavement condition grade. If the model instead always returns the prevalent class, it attains valid marginal coverage with singleton sets while providing little information about the rare non-compliant instances that an audit is intended to identify, a failure that marginal coverage alone cannot detect. Sufficient minority-class representation in the calibration set is therefore a precondition for a meaningful evaluation of these attributes, not merely a statistical convenience.

Continuous attributes such as pothole depth or curb height would use the conformalized sampling quantiles of Section~\ref{sec:continuous} unchanged. For these attributes, the main limitation is likely the imagery itself. Pedestrian-level imagery encodes metric depth and height only weakly, and the slope results show what follows, as the attribute least expressed in the image, cross slope, produced valid but uninformative intervals. Attributes that require fine three-dimensional geometry may inherit the same regime unless the imagery is complemented by depth-capable sensing. Whether a given attribute provides sufficient precision for a specific assessment remains testable from calibration data by comparing its calibrated half-width with the relevant regulatory scale and the observed field distribution.
Across both families, the bottleneck is thus empirical rather than methodological, namely sufficient label diversity in the calibration data, a model that samples the true value with non-trivial frequency, and imagery that actually expresses the attribute.

\subsection{Limitations}\label{sec:limitations}

Limitations should be considered when interpreting these results. Although overall empirical coverage is controlled, the guarantee is marginal rather than conditional. Reliability may therefore differ across urban environments, lighting conditions, or other subgroups underrepresented in the calibration data. The dispersion diagnostics make this concern concrete: coverage on the images where the models are most self-consistent is maintained only through the uniform conformal correction. Relatedly, the guarantee assumes exchangeability between calibration and deployment imagery, and two departures matter here. Our images were captured under a standardized acquisition protocol, whereas citizen-submitted imagery varies in height, angle, and quality. Recalibration on deployment-representative imagery would therefore be required before the guarantees transfer to the triage setting of Section~\ref{sec:intro}. In addition, consecutive images along the same sidewalk are spatially autocorrelated, and the random calibration--test split places images from the same segments on both sides of the split. The reported coverage is likely optimistic about transfer to unvisited sites. An area-stratified split would provide a more conservative evaluation and is recommended for future work.

Furthermore, the evaluation rests on 514 images collected in Seoul and annotated according to Korean accessibility regulations. The framework itself is independent of any particular standard, but the calibrated intervals and conclusions about their practical informativeness do not transfer to cities with different infrastructure or accessibility requirements without locally representative ground truth. The field distribution also constrains what the pavement condition grade results can establish, since with 80.2\% of images at grade A and none at grade E, the degeneracy documented in Section~\ref{sec:res-surface} reflects the scarcity of damaged pavement as much as model behavior. Moreover, pavement condition grades were assigned by a single rater without inter-rater reliability assessment, and given the inherent subjectivity of the A--E service-condition scale, this represents a source of label noise that the evaluation does not capture.

Finally, the findings are specific to the model versions, serving configurations, prompt formulation, and sampling budget evaluated. The exclusion of Claude Opus 4.6, whose deterministic decoding removed the sampling signal the framework requires (Section~\ref{sec:model-selection}), rests on a pilot of ten images. The invariance was observed on every pilot query, but the sample is too small to characterize the model across the full diversity of scenes and attributes, and the finding should be read as evidence of a serving layer configuration rather than a property of the model family. More broadly, it illustrates that provider-side changes can alter not only accuracy but the very applicability of sampling-based conformal methods.
The prompt is also part of the evaluated prediction system. In the pilot, alternative prompt formulations shifted the model responses while largely preserving their distributional shape. CP can provide coverage guarantees for any fixed prompt because the calibration data capture its associated error distribution. However, interval widths and systematic biases depend jointly on the model and prompt. Changing the prompt would therefore require recalibration. The paved-span reading of the width bias in Section~\ref{sec:synthesis} is also a testable instance, since a prompt defining width as the unobstructed passage should shift the responses downward if the reading is correct. Furthermore, the sampling budget of $M = 30$ bounds the resolution of the frequency-based scores, and conclusions about set sizes near the degeneracy boundary may shift under a larger budget.

\section{Conclusion}\label{sec:conclusion}

This paper investigated whether current VLMs can provide statistically valid and sufficiently precise estimates of sidewalk accessibility attributes from pedestrian-level imagery. Evaluating four VLMs on 514 images paired with field-measured ground truth in Seoul, we find that conformal calibration achieves the target 90\% coverage across all models and attributes, but that statistical validity does not imply sufficient precision for regulatory assessment. None of the quantitative attributes evaluated achieves sufficient precision for general regulatory compliance assessment at this coverage level. Effective width provides the most informative estimates, but even the best-performing model has a mean interval half-width of approximately 1.0~m, which remains substantial relative to the 1.5~m and 2.0~m regulatory thresholds and limits threshold determination to observations sufficiently far from those boundaries. Longitudinal slope is substantially less informative, with calibrated half-widths of the same order as the regulatory limits themselves, while cross-slope intervals are too wide to resolve the regulatory range. Pavement-condition assessment exhibits a different limitation: for three of the four models, calibrated prediction sets degenerate to all five grades because of systematic one-notch misclassification.

The study extends sampling-based CP to VLM-based sidewalk accessibility assessment across heterogeneous output types validated against field measurements. It shows that per-tail asymmetric calibration can translate directional model bias into intervals up to 33\% shorter at unchanged coverage, while also revealing when calibrated predictions remain insufficiently informative for the intended decision. This distinction is particularly important because raw sampling dispersion provides a misleading indication of uncertainty: uncalibrated intervals cover only 17--47\% of field measurements at a nominal 90\% level, and even highly self-consistent responses can systematically miss the ground truth. Response consistency therefore provides no evidence of correctness, and calibration against field-measured ground truth is a prerequisite for interpreting repeated VLM responses as uncertainty estimates in compliance-oriented applications. CP consequently provides a principled way to evaluate whether VLM uncertainty is compatible with a given assessment task, revealing from calibration data when statistically valid predictions remain too imprecise for regulatory use. To support further work, we release the 514 annotated images with their field measurements.

Taken together, these findings reframe the question for municipalities and researchers from which VLM performs best to whether the uncertainty of its predictions is sufficiently small for the specific accessibility attribute and decision under consideration. Under the conditions evaluated here, current VLMs operating on pedestrian-level imagery have not yet reached the precision required for general quantitative sidewalk compliance assessment. More broadly, sampling-based CP provides a way to move beyond unquantified VLM point predictions toward evidence-based assessment of model readiness, distinguishing statistical validity from the level of precision required by the physical and regulatory scale of a specific decision.

\section*{Acknowledgments}
The authors thank Fábio Duarte for his valuable feedback and critical review of this manuscript. The authors would also like to thank the Seoul AI Foundation and all members of the MIT Senseable City Consortium (including Seoul National University, KACST - King Abdulaziz City for Science and Technology, SMART - Singapore-MIT Alliance for Research and Technology, AMS Institute, UnipolTech, FAE Technology, Dubai Future Foundation, Sondotécnica, Arnold Ventures, Woven by Toyota, Abu Dhabi’s Department of Municipalities and Transport, Sidara, A2A, Laval, Rio de Janeiro, and Amsterdam) for supporting this study.

\section*{Data Availability}

The dataset is available at \url{https://doi.org/10.5281/zenodo.22699523}.

\bibliographystyle{unsrtnat}
\bibliography{references}

@article{Froehlich2019accessible,
author = {Froehlich, Jon E. and Brock, Anke M. and Caspi, Anat and Guerreiro, Jo\~{a}o and Hara, Kotaro and Kirkham, Reuben and Sch\"{o}ning, Johannes and Tannert, Benjamin},
title = {Grand challenges in accessible maps},
year = {2019},
issue_date = {March - April 2019},
publisher = {Association for Computing Machinery},
address = {New York, NY, USA},
volume = {26},
number = {2},
issn = {1072-5520},
url = {https://doi.org/10.1145/3301657},
doi = {10.1145/3301657},
journal = {Interactions},
month = feb,
pages = {78–81},
numpages = {4}
}

@article{Wagner2025ADA,
author = {Molly Wagner and Manish Shirgaokar and Aditi Misra and Wesley Marshall},
title = {Navigating ADA Compliance},
journal = {Journal of the American Planning Association},
volume = {91},
number = {2},
pages = {207--224},
year = {2025},
publisher = {Routledge},
doi = {10.1080/01944363.2024.2343661},
URL = {        https://doi.org/10.1080/01944363.2024.2343661
},
eprint = {         https://doi.org/10.1080/01944363.2024.2343661
}}

@misc{ada2010standards,
  title        = {2010 {ADA} Standards for Accessible Design},
  author       = {{U.S. Department of Justice}},
  year         = {2010},
  note         = {Revised 2010 ADA Standards for Accessible Design}
}

@article{Eisenberg2020barrier,
title = {Are communities in the United States planning for pedestrians with disabilities? Findings from a systematic evaluation of local government barrier removal plans},
journal = {Cities},
volume = {102},
pages = {102720},
year = {2020},
issn = {0264-2751},
doi = {10.1016/j.cities.2020.102720},
url = {https://www.sciencedirect.com/science/article/pii/S0264275119302501},
author = {Yochai Eisenberg and Amy Heider and Rob Gould and Robin Jones},
}

@Article{eisenberg2024ADAmetrics,
AUTHOR = {Eisenberg, Yochai and Hayes, Mackenzie and Hofstra, Amy and Labbé, Delphine and Gould, Robert and Jones, Robin},
TITLE = {Performance Metrics for Implementation of Americans with Disabilities Act Transition Plans},
JOURNAL = {Urban Science},
VOLUME = {8},
YEAR = {2024},
NUMBER = {2},
ARTICLE-NUMBER = {27},
URL = {https://www.mdpi.com/2413-8851/8/2/27},
ISSN = {2413-8851},
DOI = {10.3390/urbansci8020027}
}

@article{browson2009measuring,
title = {Measuring the Built Environment for Physical Activity: State of the Science},
journal = {American Journal of Preventive Medicine},
volume = {36},
number = {4, Supplement },
pages = {S99-S123.e12},
year = {2009},
note = {Measurement of the Food and Physical Activity Environments},
issn = {0749-3797},
doi = {10.1016/j.amepre.2009.01.005},
url = {https://www.sciencedirect.com/science/article/pii/S0749379709000130},
author = {Ross C. Brownson and Christine M. Hoehner and Kristen Day and Ann Forsyth and James F. Sallis},
}

@Inbook{Hartmann2017potential311,
author="Hartmann, Sarah
and Mainka, Agnes
and Stock, Wolfgang G.",
editor="Paulin, Alois A.
and Anthopoulos, Leonidas G.
and Reddick, Christopher G.",
title="Citizen Relationship Management in Local Governments: The Potential of 311 for Public Service Delivery",
bookTitle="Beyond Bureaucracy: Towards Sustainable Governance Informatisation",
year="2017",
publisher="Springer International Publishing",
address="Cham",
pages="337--353",
isbn="978-3-319-54142-6",
doi="10.1007/978-3-319-54142-6_18",
url="https://doi.org/10.1007/978-3-319-54142-6_18"
}

@article{stowers2022city311,
  title={Back to basics: City services and 311 service requests},
  author={Stowers, Genie N.L.},
  journal={State and Local Government Review},
  volume={54},
  number={1},
  pages={13--31},
  year={2022},
  publisher={SAGE Publications Sage CA: Los Angeles, CA}
}

@article{xu2017predict311,
title = {Predicting demand for 311 non-emergency municipal services: An adaptive space-time kernel approach},
journal = {Applied Geography},
volume = {89},
pages = {133-141},
year = {2017},
issn = {0143-6228},
doi = {10.1016/j.apgeog.2017.10.012},
url = {https://www.sciencedirect.com/science/article/pii/S0143622817304538},
author = {Li Xu and Mei-Po Kwan and Sara McLafferty and Shaowen Wang},
}

@article{falco2019digital,
  title={Digital participatory platforms for co-production in urban development: A systematic review},
  author={Falco, Enzo and Kleinhans, Reinout},
  journal={Crowdsourcing: Concepts, methodologies, tools, and applications},
  pages={663--690},
  year={2019},
  publisher={IGI Global Scientific Publishing}
}

@article{helbing2024co,
  title={Co-creating the future: participatory cities and digital governance},
  author={Helbing, Dirk and Mahajan, Sachit and Carpentras, Dino and Menendez, Monica and Pournaras, Evangelos and Thurner, Stefan and Verma, Trivik and Arcaute, Elsa and Batty, Michael and Bettencourt, Luis MA},
  journal={Philosophical transactions. Series A, Mathematical, physical, and engineering sciences},
  volume={382},
  number={2285},
  pages={20240113},
  year={2024}
}

@article{mclafferty2020bias311,
title = {Placing volunteered geographic health information: Socio-spatial bias in 311 bed bug report data for New York City},
journal = {Health \& Place},
volume = {62},
pages = {102282},
year = {2020},
issn = {1353-8292},
doi = {10.1016/j.healthplace.2019.102282},
url = {https://www.sciencedirect.com/science/article/pii/S1353829219309050},
author = {Sara McLafferty and Daniel Schneider and Kathryn Abelt},
}

@article{chen2024howfar,
  title   = {How far are we to {GPT-4V}? Closing the gap to commercial multimodal models with open-source suites},
  author  = {Chen, Zhe and Wang, Weiyun and Tian, Hao and Ye, Shenglong and Gao, Zhangwei and Cui, Erfei and Tong, Wenwen and Hu, Kongzhi and Luo, Jiapeng and Ma, Zheng and others},
  journal = {Science China Information Sciences},
  volume  = {67},
  number  = {12},
  pages   = {220101},
  year    = {2024},
  doi     = {10.1007/s11432-024-4231-5}
}

@article{arya2021roaddamage,
title = {Deep learning-based road damage detection and classification for multiple countries},
journal = {Automation in Construction},
volume = {132},
pages = {103935},
year = {2021},
issn = {0926-5805},
doi = {10.1016/j.autcon.2021.103935},
url = {https://www.sciencedirect.com/science/article/pii/S0926580521003861},
author = {Deeksha Arya and Hiroya Maeda and Sanjay Kumar Ghosh and Durga Toshniwal and Alexander Mraz and Takehiro Kashiyama and Yoshihide Sekimoto},
}

@article{ferrerfont2026scalablesidewalk,
author = {Pau Ferrer-Font and Istiakur Rahman and Cristina Torres-Machi},
title ={Exploring the Potential of Remote Sensing and Machine Learning for Scalable Sidewalk Condition Assessment},
journal = {Transportation Research Record},
volume = {0},
number = {0},
pages = {03611981261459434},
year = {2026},
doi = {10.1177/03611981261459434},
URL = {   https://doi.org/10.1177/03611981261459434
},
eprint = {      https://doi.org/10.1177/03611981261459434
},}

@misc{toronto2021accessibility,
  author       = {{City of Toronto}},
  title        = {Toronto Accessibility Design Guidelines},
  year         = {2021},
  howpublished = {Toronto, ON, Canada: City of Toronto}
}

@misc{usaccessboard2023prowag,
  author       = {{US Access Board}},
  title        = {Accessibility Guidelines for Pedestrian Facilities in the
                  Public Right-of-Way},
  year         = {2023},
  howpublished = {Final rule, 36 C.F.R. Part 1190, Federal Register, 8 August 2023},
  url          = {https://www.access-board.gov/prowag/}
}

@inproceedings{saha2019projectsidewalk,
author = {Saha, Manaswi and Saugstad, Michael and Maddali, Hanuma Teja and Zeng, Aileen and Holland, Ryan and Bower, Steven and Dash, Aditya and Chen, Sage and Li, Anthony and Hara, Kotaro and Froehlich, Jon},
title = {Project Sidewalk: A Web-based Crowdsourcing Tool for Collecting Sidewalk Accessibility Data At Scale},
year = {2019},
isbn = {9781450359702},
publisher = {Association for Computing Machinery},
address = {New York, NY, USA},
url = {https://doi.org/10.1145/3290605.3300292},
doi = {10.1145/3290605.3300292},
booktitle = {Proceedings of the 2019 CHI Conference on Human Factors in Computing Systems},
pages = {1–14},
numpages = {14},
location = {Glasgow, Scotland Uk},
series = {CHI '19}
}

@article{peng2025vision,
  title={Vision language model ({VLM})-enabled street view analytics: a systematic literature review},
  author={Peng, Ziyu and Lu, Weisheng and An, Hongda and Xia, Xianhua and Zhang, Yi and Xue, Fan and Chen, Junjie},
  journal={Engineering, Construction and Architectural Management},
  pages={1--19},
  year={2025},
  publisher={Emerald Publishing Limited}
}

@misc{zhang2023oasis,
      title={OASIS: Automated Assessment of Urban Pedestrian Paths at Scale}, 
      author={Yuxiang Zhang and Suresh Devalapalli and Sachin Mehta and Anat Caspi},
      year={2023},
      eprint={2303.02287},
      archivePrefix={arXiv},
      primaryClass={cs.CV},
      url={https://arxiv.org/abs/2303.02287}, 
}

@article{Jang02012025,
author = {Kee Moon Jang and Junghwan Kim},
title = {Multimodal Large Language Models as Built Environment Auditing Tools},
journal = {The Professional Geographer},
volume = {77},
number = {1},
pages = {84--90},
year = {2025},
publisher = {Routledge},
doi = {10.1080/00330124.2024.2404894},
URL = {      https://doi.org/10.1080/00330124.2024.2404894
},
eprint = {        https://doi.org/10.1080/00330124.2024.2404894
}
}

@misc{wang2026vlmssensorsfeelscalable,
      title={Do {VLMs} See What Sensors Feel? A Scalable Expert-Guided Design for Wheelchair Accessibility Assessment from Street View}, 
      author={Dongdong Wang and Alina Hagen and Isabelle Gatmaitan and Hao Zhou and Yiwen Dong and Shabboo Valipoor and Vivian W. H. Wong and Lingyao Li},
      year={2026},
      eprint={2606.07642},
      archivePrefix={arXiv},
      primaryClass={cs.CV},
      url={https://arxiv.org/abs/2606.07642}, 
}

@misc{lalwani2026VLMannotation,
      title={Towards Human-{AI} Accessibility Mapping in India: {VLM-}Guided Annotations and POI-Centric Analysis in Chandigarh}, 
      author={Varchita Lalwani and Utkarsh Agarwal and Michael Saugstad and Manish Kumar and Jon E. Froehlich and Anupam Sobti},
      year={2026},
      eprint={2602.09216},
      archivePrefix={arXiv},
      primaryClass={cs.HC},
      url={https://arxiv.org/abs/2602.09216}, 
}

@misc{kostumov2024uncertaintyaware,
  title         = {Uncertainty-Aware Evaluation for Vision-Language Models},
  author        = {Kostumov, Vasily and Nutfullin, Bulat and Pilipenko, Oleg
                   and Ilyushin, Eugene},
  year          = {2024},
  eprint        = {2402.14418},
  archivePrefix = {arXiv},
  primaryClass  = {cs.CV},
  url           = {https://arxiv.org/abs/2402.14418}
}

@inproceedings{groot2024overconfidence,
    title = "Overconfidence is Key: Verbalized Uncertainty Evaluation in Large Language and Vision-Language Models",
    author = "Groot, Tobias  and
      Valdenegro - Toro, Matias",
    editor = "Ovalle, Anaelia  and
      Chang, Kai-Wei  and
      Cao, Yang Trista  and
      Mehrabi, Ninareh  and
      Zhao, Jieyu  and
      Galstyan, Aram  and
      Dhamala, Jwala  and
      Kumar, Anoop  and
      Gupta, Rahul",
    booktitle = "Proceedings of the 4th Workshop on Trustworthy Natural Language Processing (TrustNLP 2024)",
    month = jun,
    year = "2024",
    address = "Mexico City, Mexico",
    publisher = "Association for Computational Linguistics",
    url = "https://aclanthology.org/2024.trustnlp-1.13/",
    doi = "10.18653/v1/2024.trustnlp-1.13",
    pages = "145--171",
}

@inproceedings{xiong2024LLMUuncertainty,
 author = {Xiong, Miao and Hu, Zhiyuan and Lu, Xinyang and LI, YIFEI and Fu, Jie and He, Junxian and Hooi, Bryan},
 booktitle = {International Conference on Learning Representations},
 editor = {B. Kim and Y. Yue and S. Chaudhuri and K. Fragkiadaki and M. Khan and Y. Sun},
 pages = {23650--23678},
 title = {Can {LLMs} Express Their Uncertainty? An Empirical Evaluation of Confidence Elicitation in {LLMs}},
 url = {https://proceedings.iclr.cc/paper_files/paper/2024/file/6733cf15e10e2cd1d59af033c3bb8507-Paper-Conference.pdf},
 volume = {2024},
 year = {2024}
}

@book{vovk2005algorithmic,
  title={Algorithmic learning in a random world},
  author={Vovk, Vladimir and Gammerman, Alexander and Shafer, Glenn},
  year={2005},
  publisher={Springer}
}

@article{angelopoulos2023conformalprediction,
author = {Angelopoulos, Anastasios N. and Bates, Stephen},
title = {Conformal Prediction: A Gentle Introduction},
year = {2023},
issue_date = {Mar 2023},
publisher = {Now Publishers Inc.},
address = {Hanover, MA, USA},
volume = {16},
number = {4},
issn = {1935-8237},
url = {https://doi.org/10.1561/2200000101},
doi = {10.1561/2200000101},
journal = {Found. Trends Mach. Learn.},
month = mar,
pages = {494–591},
numpages = {114}
}

@article{bai2025qwen3vl,
  title   = {Qwen3-VL Technical Report},
  author  = {Bai, Shuai and Cai, Yuxuan and Chen, Ruizhe and Chen, Keqin and Chen, Xionghui and Cheng, Zesen and Deng, Lianghao and Ding, Wei and Gao, Chang and Ge, Chunjiang and others},
  journal = {arXiv preprint arXiv:2511.21631},
  year    = {2025}
}

@article{wang2025internvl35,
  title   = {InternVL3.5: Advancing Open-Source Multimodal Models in Versatility, Reasoning, and Efficiency},
  author  = {Wang, Weiyun and Gao, Zhangwei and Gu, Lixin and Pu, Hengjun and Cui, Long and Wei, Xingguang and Liu, Zhaoyang and Jing, Linglin and Ye, Shenglong and Shao, Jie and others},
  journal = {arXiv preprint arXiv:2508.18265},
  year    = {2025}
}

@article{campos2024conforlamNL,
    title = "Conformal Prediction for Natural Language Processing: A Survey",
    author = "Campos, Margarida  and
      Farinhas, Ant{\'o}nio  and
      Zerva, Chrysoula  and
      Figueiredo, M{\'a}rio A. T.  and
      Martins, Andr{\'e} F. T.",
    journal = "Transactions of the Association for Computational Linguistics",
    volume = "12",
    year = "2024",
    address = "Cambridge, MA",
    publisher = "MIT Press",
    url = "https://aclanthology.org/2024.tacl-1.82/",
    doi = "10.1162/tacl_a_00715",
    pages = "1497--1516",
}

@inproceedings{su2024APIconformal,
    title = "{API} Is Enough: Conformal Prediction for Large Language Models Without Logit-Access",
    author = "Su, Jiayuan  and
      Luo, Jing  and
      Wang, Hongwei  and
      Cheng, Lu",
    editor = "Al-Onaizan, Yaser  and
      Bansal, Mohit  and
      Chen, Yun-Nung",
    booktitle = "Findings of the Association for Computational Linguistics: EMNLP 2024",
    month = nov,
    year = "2024",
    address = "Miami, Florida, USA",
    publisher = "Association for Computational Linguistics",
    url = "https://aclanthology.org/2024.findings-emnlp.54/",
    doi = "10.18653/v1/2024.findings-emnlp.54",
    pages = "979--995",
}

@Article{hang2025neuralradiance,
AUTHOR = {Du, Hang and Wang, Shuaizhou and Zhang, Linlin and Amo-Boateng, Mark and Adu-Gyamfi, Yaw},
TITLE = {Evaluating Neural Radiance Fields for {ADA}-Compliant Sidewalk Assessments: A Comparative Study with LiDAR and Manual Methods},
JOURNAL = {Infrastructures},
VOLUME = {10},
YEAR = {2025},
NUMBER = {8},
ARTICLE-NUMBER = {191},
URL = {https://www.mdpi.com/2412-3811/10/8/191},
ISSN = {2412-3811},
DOI = {10.3390/infrastructures10080191}
}

@Article{lee2026lidarSidewalks,
AUTHOR = {Lee, Dongha and Kang, Sungho and Lee, Jaecheol and Kim, Junghyun},
TITLE = {Comparative Performance Evaluation of Multi-Type {LiDAR} Sensors and Their Applicability to Sidewalk {HD} Mapping},
JOURNAL = {Sensors},
VOLUME = {26},
YEAR = {2026},
NUMBER = {5},
ARTICLE-NUMBER = {1480},
URL = {https://www.mdpi.com/1424-8220/26/5/1480},
PubMedID = {41829444},
ISSN = {1424-8220},
DOI = {10.3390/s26051480}
}

@inproceedings{weld2019sidewalkstreetscape,
author = {Weld, Galen and Jang, Esther and Li, Anthony and Zeng, Aileen and Heimerl, Kurtis and Froehlich, Jon E.},
title = {Deep Learning for Automatically Detecting Sidewalk Accessibility Problems Using Streetscape Imagery},
year = {2019},
isbn = {9781450366762},
publisher = {Association for Computing Machinery},
address = {New York, NY, USA},
url = {https://doi.org/10.1145/3308561.3353798},
doi = {10.1145/3308561.3353798},
booktitle = {Proceedings of the 21st International ACM SIGACCESS Conference on Computers and Accessibility},
pages = {196–209},
numpages = {14},
location = {Pittsburgh, PA, USA},
series = {ASSETS '19}
}

@inproceedings{duan2022crowdAI,
author = {Duan, Michael and Kiami, Shosuke and Milandin, Logan and Kuang, Johnson and Saugstad, Michael and Hosseini, Maryam and Froehlich, Jon E.},
title = {Scaling Crowd+{AI} Sidewalk Accessibility Assessments: Initial Experiments Examining Label Quality and Cross-city Training on Performance},
year = {2022},
isbn = {9781450392587},
publisher = {Association for Computing Machinery},
address = {New York, NY, USA},
url = {https://doi.org/10.1145/3517428.3550381},
doi = {10.1145/3517428.3550381},
booktitle = {Proceedings of the 24th International ACM SIGACCESS Conference on Computers and Accessibility},
articleno = {82},
numpages = {5},
location = {Athens, Greece},
series = {ASSETS '22}
}

@inproceedings{liu2024finegrainedsidewalk,
author = {Liu, Xinlei and Wu, Kevin and Kulkarni, Minchu and Saugstad, Michael and Rapo, Peyton Anton and Freiburger, Jeremy and Hosseini, Maryam and Li, Chu and Froehlich, Jon E.},
title = {Towards Fine-Grained Sidewalk Accessibility Assessment with Deep Learning: Initial Benchmarks and an Open Dataset},
year = {2024},
isbn = {9798400706776},
publisher = {Association for Computing Machinery},
address = {New York, NY, USA},
url = {https://doi.org/10.1145/3663548.3688531},
doi = {10.1145/3663548.3688531},
booktitle = {Proceedings of the 26th International ACM SIGACCESS Conference on Computers and Accessibility},
articleno = {103},
numpages = {12},
location = {St. John's, NL, Canada},
series = {ASSETS '24}
}

@article{hosseini2023aerial,
title = {Mapping the walk: A scalable computer vision approach for generating sidewalk network datasets from aerial imagery},
journal = {Computers, Environment and Urban Systems},
volume = {101},
pages = {101950},
year = {2023},
issn = {0198-9715},
doi = {10.1016/j.compenvurbsys.2023.101950},
url = {https://www.sciencedirect.com/science/article/pii/S0198971523000133},
author = {Maryam Hosseini and Andres Sevtsuk and Fabio Miranda and Roberto M. Cesar and Claudio T. Silva},
}

@article{hou2020networklevel,
title = {A network-level sidewalk inventory method using mobile LiDAR and deep learning},
journal = {Transportation Research Part C: Emerging Technologies},
volume = {119},
pages = {102772},
year = {2020},
issn = {0968-090X},
doi = {10.1016/j.trc.2020.102772},
url = {https://www.sciencedirect.com/science/article/pii/S0968090X20306823},
author = {Qing Hou and Chengbo Ai},
}

@InProceedings{blecic2024urbanwalkability,
author="Ble{\v{c}}i{\'{c}}, Ivan
and Saiu, Valeria
and A. Trunfio, Giuseppe",
editor="Gervasi, Osvaldo
and Murgante, Beniamino
and Garau, Chiara
and Taniar, David
and C. Rocha, Ana Maria A.
and Faginas Lago, Maria Noelia",
title="Enhancing Urban Walkability Assessment with Multimodal Large Language Models",
booktitle="Computational Science and Its Applications -- ICCSA 2024 Workshops",
year="2024",
publisher="Springer Nature Switzerland",
address="Cham",
pages="394--411",
isbn="978-3-031-65282-0"
}

@misc{cai2025LLMurban,
      title={Can a Large Language Model Assess Urban Design Quality? Evaluating Walkability Metrics Across Expertise Levels}, 
      author={Chenyi Cai and Kosuke Kuriyama and Youlong Gu and Filip Biljecki and Pieter Herthogs},
      year={2025},
      eprint={2504.21040},
      archivePrefix={arXiv},
      primaryClass={cs.CV},
      url={https://arxiv.org/abs/2504.21040}, 
}

@misc{tan2026urbanvggt,
      title={UrbanVGGT: Scalable Sidewalk Width Estimation from Street View Images}, 
      author={Kaizhen Tan and Fan Zhang},
      year={2026},
      eprint={2603.22531},
      archivePrefix={arXiv},
      primaryClass={cs.CV},
      url={https://arxiv.org/abs/2603.22531}, 
}

@article{ki2023humancentric,
title = {Bridging the gap between pedestrian and street views for human-centric environment measurement: A GIS-based 3D virtual environment},
journal = {Landscape and Urban Planning},
volume = {240},
pages = {104873},
year = {2023},
issn = {0169-2046},
doi = {10.1016/j.landurbplan.2023.104873},
url = {https://www.sciencedirect.com/science/article/pii/S0169204623001925},
author = {Donghwan Ki and Keundeok Park and Zhenhua Chen},
}

@article{ito2024streetview,
author = {Koichi Ito and Matias Quintana and Xianjing Han and Roger Zimmermann and Filip Biljecki},
title = {Translating street view imagery to correct perspectives to enhance bikeability and walkability studies},
journal = {International Journal of Geographical Information Science},
volume = {38},
number = {12},
pages = {2514--2544},
year = {2024},
publisher = {Taylor \& Francis},
doi = {10.1080/13658816.2024.2391969},
URL = { 
        https://doi.org/10.1080/13658816.2024.2391969
},
eprint = {   
        https://doi.org/10.1080/13658816.2024.2391969
}
}

@article{perez2025generativeAI,
title = {Streetscape Analysis with Generative AI (SAGAI): Vision-language assessment and mapping of urban scenes},
journal = {Geomatica},
volume = {77},
number = {2},
pages = {100063},
year = {2025},
issn = {1195-1036},
doi = {10.1016/j.geomat.2025.100063},
url = {https://www.sciencedirect.com/science/article/pii/S1195103625000199},
author = {Joan Perez and Giovanni Fusco},
}

@article{Dai2024systematicreview,
author = {Shaoqing Dai and Yuchen Li and Alfred Stein and Shujuan Yang and Peng Jia},
title = {Street view imagery-based built environment auditing tools: a systematic review},
journal = {International Journal of Geographical Information Science},
volume = {38},
number = {6},
pages = {1136--1157},
year = {2024},
publisher = {Taylor \& Francis},
doi = {10.1080/13658816.2024.2336034},
URL = {   
        https://doi.org/10.1080/13658816.2024.2336034
},
eprint = {  
        https://doi.org/10.1080/13658816.2024.2336034
}}

@misc{atil2025nondeterminism,
      title={Non-Determinism of "Deterministic" {LLM} Settings}, 
      author={Berk Atil and Sarp Aykent and Alexa Chittams and Lisheng Fu and Rebecca J. Passonneau and Evan Radcliffe and Guru Rajan Rajagopal and Adam Sloan and Tomasz Tudrej and Ferhan Ture and Zhe Wu and Lixinyu Xu and Breck Baldwin},
      year={2025},
      eprint={2408.04667},
      archivePrefix={arXiv},
      primaryClass={cs.CL},
      url={https://arxiv.org/abs/2408.04667}, 
}

@inproceedings{sclar2024quantifyllm,
 author = {Sclar, Melanie and Choi, Yejin and Tsvetkov, Yulia and Suhr, Alane},
 booktitle = {International Conference on Learning Representations},
 editor = {B. Kim and Y. Yue and S. Chaudhuri and K. Fragkiadaki and M. Khan and Y. Sun},
 pages = {25055--25083},
 title = {Quantifying Language Models\textquotesingle  Sensitivity to Spurious Features in Prompt Design or: How I learned to start worrying about prompt formatting},
 url = {https://proceedings.iclr.cc/paper_files/paper/2024/file/6c0e99d736da621403018ca7b32b1a4d-Paper-Conference.pdf},
 volume = {2024},
 year = {2024}
}

@inproceedings{tian2023justask,
    title = "Just Ask for Calibration: Strategies for Eliciting Calibrated Confidence Scores from Language Models Fine-Tuned with Human Feedback",
    author = "Tian, Katherine  and
      Mitchell, Eric  and
      Zhou, Allan  and
      Sharma, Archit  and
      Rafailov, Rafael  and
      Yao, Huaxiu  and
      Finn, Chelsea  and
      Manning, Christopher",
    editor = "Bouamor, Houda  and
      Pino, Juan  and
      Bali, Kalika",
    booktitle = "Proceedings of the 2023 Conference on Empirical Methods in Natural Language Processing",
    month = dec,
    year = "2023",
    address = "Singapore",
    publisher = "Association for Computational Linguistics",
    url = "https://aclanthology.org/2023.emnlp-main.330/",
    doi = "10.18653/v1/2023.emnlp-main.330",
    pages = "5433--5442",
}

@InProceedings{guo2017calibration,
  title = 	 {On Calibration of Modern Neural Networks},
  author =       {Chuan Guo and Geoff Pleiss and Yu Sun and Kilian Q. Weinberger},
  booktitle = 	 {Proceedings of the 34th International Conference on Machine Learning},
  pages = 	 {1321--1330},
  year = 	 {2017},
  editor = 	 {Precup, Doina and Teh, Yee Whye},
  volume = 	 {70},
  series = 	 {Proceedings of Machine Learning Research},
  month = 	 {06--11 Aug},
  publisher =    {PMLR},
  url = 	 {https://proceedings.mlr.press/v70/guo17a.html},
}

@inproceedings{geng2024survey,
    title = "A Survey of Confidence Estimation and Calibration in Large Language Models",
    author = "Geng, Jiahui  and
      Cai, Fengyu  and
      Wang, Yuxia  and
      Koeppl, Heinz  and
      Nakov, Preslav  and
      Gurevych, Iryna",
    editor = "Duh, Kevin  and
      Gomez, Helena  and
      Bethard, Steven",
    booktitle = "Proceedings of the 2024 Conference of the North American Chapter of the Association for Computational Linguistics: Human Language Technologies (Volume 1: Long Papers)",
    month = jun,
    year = "2024",
    address = "Mexico City, Mexico",
    publisher = "Association for Computational Linguistics",
    url = "https://aclanthology.org/2024.naacl-long.366/",
    doi = "10.18653/v1/2024.naacl-long.366",
    pages = "6577--6595",
}

@misc{wang2023selfconsistency,
      title={Self-Consistency Improves Chain of Thought Reasoning in Language Models}, 
      author={Xuezhi Wang and Jason Wei and Dale Schuurmans and Quoc Le and Ed Chi and Sharan Narang and Aakanksha Chowdhery and Denny Zhou},
      year={2023},
      eprint={2203.11171},
      archivePrefix={arXiv},
      primaryClass={cs.CL},
      url={https://arxiv.org/abs/2203.11171}, 
}

@article{shorinwa2025survey,
author = {Shorinwa, Ola and Mei, Zhiting and Lidard, Justin and Ren, Allen Z. and Majumdar, Anirudha},
title = {A Survey on Uncertainty Quantification of Large Language Models: Taxonomy, Open Research Challenges, and Future Directions},
year = {2025},
issue_date = {February 2026},
publisher = {Association for Computing Machinery},
address = {New York, NY, USA},
volume = {58},
number = {3},
issn = {0360-0300},
url = {https://doi.org/10.1145/3744238},
doi = {10.1145/3744238},
journal = {ACM Comput. Surv.},
month = sep,
articleno = {63},
numpages = {38}
}

@misc{zhang2024vluncertaintydetecting,
title={VL-Uncertainty: Detecting Hallucination in Large Vision-Language Model via Uncertainty Estimation}, 
author={Ruiyang Zhang and Hu Zhang and Zhedong Zheng},
year={2024},
eprint={2411.11919},
archivePrefix={arXiv},
primaryClass={cs.CV},
url={https://arxiv.org/abs/2411.11919}, 
}

@article{savage2025large,
  title={Large language model uncertainty proxies: discrimination and calibration for medical diagnosis and treatment},
  author={Savage, Thomas and Wang, John and Gallo, Robert and Boukil, Abdessalem and Patel, Vishwesh and Safavi-Naini, Seyed Amir Ahmad and Soroush, Ali and Chen, Jonathan H},
  journal={Journal of the American Medical Informatics Association},
  volume={32},
  number={1},
  pages={139--149},
  year={2025},
  publisher={Oxford University Press}
}

@inproceedings{tan2025consistent,
    title = "Too Consistent to Detect: A Study of Self-Consistent Errors in {LLM}s",
    author = "Tan, Hexiang  and
      Sun, Fei  and
      Liu, Sha  and
      Su, Du  and
      Cao, Qi  and
      Chen, Xin  and
      Wang, Jingang  and
      Cai, Xunliang  and
      Wang, Yuanzhuo  and
      Shen, Huawei  and
      Cheng, Xueqi",
    editor = "Christodoulopoulos, Christos  and
      Chakraborty, Tanmoy  and
      Rose, Carolyn  and
      Peng, Violet",
    booktitle = "Proceedings of the 2025 Conference on Empirical Methods in Natural Language Processing",
    month = nov,
    year = "2025",
    address = "Suzhou, China",
    publisher = "Association for Computational Linguistics",
    url = "https://aclanthology.org/2025.emnlp-main.238/",
    doi = "10.18653/v1/2025.emnlp-main.238",
    pages = "4755--4765",
    ISBN = "979-8-89176-332-6",
}

@article{shafer2008tutorial,
  title={A tutorial on conformal prediction.},
  author={Shafer, Glenn and Vovk, Vladimir},
  journal={Journal of machine learning research},
  volume={9},
  number={3},
  year={2008}
}

@InProceedings{papadopoulos2002inductiveconfidence,
author="Papadopoulos, Harris
and Proedrou, Kostas
and Vovk, Volodya
and Gammerman, Alex",
editor="Elomaa, Tapio
and Mannila, Heikki
and Toivonen, Hannu",
title="Inductive Confidence Machines for Regression",
booktitle="Machine Learning: ECML 2002",
year="2002",
publisher="Springer Berlin Heidelberg",
address="Berlin, Heidelberg",
pages="345--356",
isbn="978-3-540-36755-0"
}

@inproceedings{romano2019quantile,
 author = {Romano, Yaniv and Patterson, Evan and Candes, Emmanuel},
 booktitle = {Advances in Neural Information Processing Systems},
 editor = {H. Wallach and H. Larochelle and A. Beygelzimer and F. d\textquotesingle Alch\'{e}-Buc and E. Fox and R. Garnett},
 pages = {},
 publisher = {Curran Associates, Inc.},
 title = {Conformalized Quantile Regression},
 url = {https://proceedings.neurips.cc/paper_files/paper/2019/file/5103c3584b063c431bd1268e9b5e76fb-Paper.pdf},
 volume = {32},
 year = {2019}
}

@article{Sadinle2019classifier,
author = {Mauricio Sadinle and Jing Lei and Larry Wasserman},
title = {Least Ambiguous Set-Valued Classifiers With Bounded Error Levels},
journal = {Journal of the American Statistical Association},
volume = {114},
number = {525},
pages = {223--234},
year = {2019},
publisher = {Taylor \& Francis},
doi = {10.1080/01621459.2017.1395341},
URL = { 
        https://doi.org/10.1080/01621459.2017.1395341
},
eprint = {    
        https://doi.org/10.1080/01621459.2017.1395341
}}

@inproceedings{romano2020neural,
 author = {Romano, Yaniv and Sesia, Matteo and Candes, Emmanuel},
 booktitle = {Advances in Neural Information Processing Systems},
 editor = {H. Larochelle and M. Ranzato and R. Hadsell and M.F. Balcan and H. Lin},
 pages = {3581--3591},
 publisher = {Curran Associates, Inc.},
 title = {Classification with Valid and Adaptive Coverage},
 url = {https://proceedings.neurips.cc/paper_files/paper/2020/file/244edd7e85dc81602b7615cd705545f5-Paper.pdf},
 volume = {33},
 year = {2020}
}

@article{zhou2025data,
author = {Zhou, Xiaofan and Chen, Baiting and Gui, Yu and Cheng, Lu},
title = {Conformal Prediction: A Data Perspective},
year = {2025},
issue_date = {January 2026},
publisher = {Association for Computing Machinery},
address = {New York, NY, USA},
volume = {58},
number = {2},
issn = {0360-0300},
url = {https://doi.org/10.1145/3736575},
doi = {10.1145/3736575},
journal = {ACM Comput. Surv.},
month = sep,
articleno = {49},
numpages = {37}
}

@inproceedings{quach2024conformal,
 author = {Quach, Victor and Fisch, Adam and Schuster, Tal and Yala, Adam and Sohn, Jae Ho and Jaakkola, Tommi and Barzilay, Regina },
 booktitle = {International Conference on Learning Representations},
 editor = {B. Kim and Y. Yue and S. Chaudhuri and K. Fragkiadaki and M. Khan and Y. Sun},
 pages = {11654--11681},
 title = {Conformal Language Modeling},
 url = {https://proceedings.iclr.cc/paper_files/paper/2024/file/31421b112e5f7faf4fc577b74e45dab2-Paper-Conference.pdf},
 volume = {2024},
 year = {2024}
}

@misc{mohri2024languagemodels,
      title={Language Models with Conformal Factuality Guarantees}, 
      author={Christopher Mohri and Tatsunori Hashimoto},
      year={2024},
      eprint={2402.10978},
      archivePrefix={arXiv},
      primaryClass={cs.LG},
      url={https://arxiv.org/abs/2402.10978}, 
}

@inproceedings{cherian2024LLMvalidity,
 author = {Cherian, John J. and Gibbs, Isaac and Cand\`{e}s, Emmanuel J.},
 booktitle = {Advances in Neural Information Processing Systems},
 doi = {10.52202/079017-3645},
 editor = {A. Globerson and L. Mackey and D. Belgrave and A. Fan and U. Paquet and J. Tomczak and C. Zhang},
 pages = {114812--114842},
 publisher = {Curran Associates, Inc.},
 title = {Large language model validity via enhanced conformal prediction methods},
 url = {https://proceedings.neurips.cc/paper_files/paper/2024/file/d02ff1aeaa5c268dc34790dd1ad21526-Paper-Conference.pdf},
 volume = {37},
 year = {2024}
}

@InProceedings{Silva2025CVPR,
    author    = {Silva-Rodr{\'\i}guez, Julio and Ben Ayed, Ismail and Dolz, Jose},
    title     = {Conformal Prediction for Zero-Shot Models},
    booktitle = {Proceedings of the IEEE/CVF Conference on Computer Vision and Pattern Recognition (CVPR)},
    month     = {June},
    year      = {2025},
    pages     = {19931-19941}
}

@InProceedings{silva2026fullconformal,
author="Silva-Rodr{\'i}guez, Julio
and Fillioux, Leo
and Courn{\`e}de, Paul-Henry
and Vakalopoulou, Maria
and Christodoulidis, Stergios
and Ayed, Ismail Ben
and Dolz, Jose",
editor="Oguz, Ipek
and Zhang, Shaoting
and Metaxas, Dimitris N.",
title="Full Conformal Adaptation of Medical Vision-Language Models",
booktitle="Information Processing in Medical Imaging",
year="2026",
publisher="Springer Nature Switzerland",
address="Cham",
pages="278--293",
isbn="978-3-031-96625-5"
}

@misc{ye2025datadrivencalibration,
      title={Data-Driven Calibration of Prediction Sets in Large Vision-Language Models Based on Inductive Conformal Prediction}, 
      author={Yuanchang Ye and Weiyan Wen},
      year={2025},
      eprint={2504.17671},
      archivePrefix={arXiv},
      primaryClass={cs.CL},
      url={https://arxiv.org/abs/2504.17671}, 
}

@misc{alyassirad2026conrep,
      title={CONRep: Uncertainty-Aware Vision-Language Report Drafting Using Conformal Prediction}, 
      author={Danial Elyassirad and Benyamin Gheiji and Mahsa Vatanparast and Amir Mahmoud Ahmadzadeh and Seyed Amir Asef Agah and Mana Moassefi and Meysam Tavakoli and Shahriar Faghani},
      year={2026},
      eprint={2602.03910},
      archivePrefix={arXiv},
      primaryClass={eess.IV},
      url={https://arxiv.org/abs/2602.03910}, 
}

@misc{epstein2025llms,
      title={{LLMs} are Overconfident: Evaluating Confidence Interval Calibration with FermiEval}, 
      author={Elliot Epstein and John Winnicki and Thanawat Sornwanee and Rajat Dwaraknath},
      year={2025},
      eprint={2510.26995},
      archivePrefix={arXiv},
      primaryClass={eess.IV},
      url={https://arxiv.org/abs/2510.26995}, 
}

@misc{tayebati2025conformalpolicies,
      title={Learning Conformal Abstention Policies for Adaptive Risk Management in Large Language and Vision-Language Models}, 
      author={Sina Tayebati and Divake Kumar and Nastaran Darabi and Dinithi Jayasuriya and Ranganath Krishnan and Amit Ranjan Trivedi},
      year={2025},
      eprint={2502.06884},
      archivePrefix={arXiv},
      primaryClass={cs.LG},
      url={https://arxiv.org/abs/2502.06884}, 
}

@article{fillioux2026foundationmodels,
title = "Are foundation models for computer vision good conformal predictors?",
author  = {Fillioux, Leo and Silva-Rodr{\'i}guez, Julio and Ben Ayed, Ismail and Courn{\`e}de, Paul-Henry and Vakalopoulou, Maria and Christodoulidis, Stergios and Dolz, Jose},
note = "Publisher Copyright: {\textcopyright} 2026, Transactions on Machine Learning Research. All rights reserved.",
year = "2026",
language = "English",
volume = "2026-March",
journal = "Transactions on Machine Learning Research",
issn = "2835-8856",
publisher = "Transactions on Machine Learning Research",
}

@INPROCEEDINGS{yang2025frequency,
  author={Yang, Guang and Zhang, YongLiang and Liu, XinYang and Wu, Zhuoqun},
  booktitle={2025 4th International Conference on Image Processing, Computer Vision and Machine Learning (ICICML)}, 
  title={Frequency-Based Predictive Entropy for Uncertainty Quantification in Black-Box Multiple-Choice Question Answering}, 
  year={2025},
  volume={},
  number={},
  pages={1782-1786},
  doi={10.1109/ICICML67980.2025.11333521}}

@misc{yang2025conformalsets,
      title={Conformal Sets in Multiple-Choice Question Answering under Black-Box Settings with Provable Coverage Guarantees}, 
      author={Guang Yang and Xinyang Liu},
      year={2025},
      eprint={2508.05544},
      archivePrefix={arXiv},
      primaryClass={cs.CL},
      url={https://arxiv.org/abs/2508.05544}, 
}

@inproceedings{chiang2024chatbotarena,
author = {Chiang, Wei-Lin and Zheng, Lianmin and Sheng, Ying and Angelopoulos, Anastasios N. and Li, Tianle and Li, Dacheng and Zhu, Banghua and Zhang, Hao and Jordan, Michael I. and Gonzalez, Joseph E. and Stoica, Ion},
title = {Chatbot arena: an open platform for evaluating {LLMs} by human preference},
year = {2024},
publisher = {JMLR.org},
booktitle = {Proceedings of the 41st International Conference on Machine Learning},
articleno = {331},
numpages = {30},
location = {Vienna, Austria},
series = {ICML'24}
}

@inproceedings{wang2025sconu,
    title = "{SC}on{U}: Selective Conformal Uncertainty in Large Language Models",
    author = "Wang, Zhiyuan  and
      Wang, Qingni  and
      Zhang, Yue  and
      Chen, Tianlong  and
      Zhu, Xiaofeng  and
      Shi, Xiaoshuang  and
      Xu, Kaidi",
    editor = "Che, Wanxiang  and
      Nabende, Joyce  and
      Shutova, Ekaterina  and
      Pilehvar, Mohammad Taher",
    booktitle = "Proceedings of the 63rd Annual Meeting of the Association for Computational Linguistics (Volume 1: Long Papers)",
    month = jul,
    year = "2025",
    address = "Vienna, Austria",
    publisher = "Association for Computational Linguistics",
    url = "https://aclanthology.org/2025.acl-long.934/",
    doi = "10.18653/v1/2025.acl-long.934",
    pages = "19052--19075",
    ISBN = "979-8-89176-251-0",
}

@misc{zeng2026empirical,
      title={Empirical Bayes Conformal Prediction for Vision and Language Models}, 
      author={Jiapeng Zeng and Yogesh Prabhu and Zhanpeng Zeng and Michael A. Newton and Vikas Singh},
      year={2026},
      eprint={2605.23189},
      archivePrefix={arXiv},
      primaryClass={cs.LG},
      url={https://arxiv.org/abs/2605.23189}, 
}

@inproceedings{lin2026domainshift,
title={Domain-Shift-Aware Conformal Prediction for Large Language Models},
author={Zhexiao Lin and Yuanyuan Li and Neeraj Sarna and Yuanyuan Gao and Michael Berger},
booktitle={Forty-third International Conference on Machine Learning},
year={2026},
url={https://openreview.net/forum?id=UbzajT84Nd}
}

@manual{molit2021sidewalk,
  author  = {{Ministry of Land, Infrastructure and Transport}},
  title   = {Guidelines for Installation and Management of Sidewalks},
  year    = {2021},
  month   = jul,
  note    = {In Korean; title translated by the authors},
  url     = {https://www.law.go.kr/%ED%96%89%EC%A0%95%EA%B7%9C%EC%B9%99/%EB%B3%B4%EB%8F%84%EC%84%A4%EC%B9%98%EB%B0%8F%EA%B4%80%EB%A6%AC%EC%A7%80%EC%B9%A8},
  urldate = {2026-08-02}
}

@manual{seoul2026sidewalk,
  author = {{Seoul Metropolitan Government, Road Management Division}},
  title  = {Seoul Sidewalk Design and Construction Manual, Version 3.0},
  year   = {2026},
  month  = jan,
  note   = {In Korean; title translated by the authors}
}

@techreport{kotsa2022pedestrian,
  author      = {{Korea Transportation Safety Authority}},
  title       = {National Survey of Pedestrian Transportation Conditions and Study on Its Utilization: Final Report},
  institution = {Ministry of Land, Infrastructure and Transport},
  year        = {2022},
  month       = dec,
  note        = {In Korean; title translated by the authors}
}

@misc{korea2025mobility,
  author  = {{Republic of Korea}},
  title   = {{Act on Promotion of the Transportation Convenience of Mobility Disadvantaged Persons}},
  year    = {2025},
  note    = {Act No. 20756, amended January 31, 2025; effective February 1, 2026},
  url     = {https://www.law.go.kr/%EB%B2%95%EB%A0%B9/%EA%B5%90%ED%86%B5%EC%95%BD%EC%9E%90%EC%9D%98%EC%9D%B4%EB%8F%99%ED%8E%B8%EC%9D%98%EC%A6%9D%EC%A7%84%EB%B2%95},
  urldate = {2026-08-02}
}

@article{ai2015automated,
author = {Chengbo Ai and Yichang (James) Tsai},
title ={Automated Sidewalk Assessment Method for Americans with Disabilities Act Compliance Using Three-Dimensional Mobile Lidar},
journal = {Transportation Research Record},
volume = {2542},
number = {1},
pages = {25-32},
year = {2016},
doi = {10.3141/2542-04},
URL = { 
        https://doi.org/10.3141/2542-04
},
eprint = { 
        https://doi.org/10.3141/2542-04
}
,
}

\appendix

\section{Alternative width prompt trialed in the pilot}
\label{app:prompt-alt}
The longer formulation below was trialed against the concise prompt of Appendix~\ref{app:prompts} during the pilot and not retained.

\begin{verbatim}
You are an expert in visual scene understanding analyzing a street-view image.
Task: Estimate the clear walking width of the sidewalk (pedestrian footpath) in meters.

Definition:
Sidewalk = the pedestrian walking surface intended for pedestrians.
Width = the distance from the inner sidewalk edge to the outer sidewalk edge,
measured perpendicular to the walking direction.
Estimate the clear usable walking width, excluding temporary obstacles such as
parked bicycles, pedestrians, signs, poles, benches, vegetation, or street
furniture.
Do NOT measure the road, parking lane, curb height, driveway width, or building
setback.

Estimation rules:
Use visible scene geometry, perspective cues, and nearby objects for scale.
If the sidewalk width varies, estimate the typical width visible in the image.
If only part of the sidewalk is visible, provide the best estimate for the full
sidewalk width.

Output rules:
Return exactly one numeric value.
Use meters as the unit.
Round to one decimal place.
Do not include units, text, ranges, confidence scores, or explanations.
Examples of valid outputs: 1.2
\end{verbatim}

\section{Attribute Prompts}
\label{app:prompts}

Each prompt was issued verbatim, one per request, together with a single image and no system prompt or
in-context examples. All four were held fixed across the four models and all 514 images.

\paragraph{Width}
{\small
\begin{verbatim}
You are an expert analyzing the provided street view image.

Estimate the WIDTH of the SIDEWALK (the pedestrian footpath), in METERS.
This means the clear horizontal walking width of the sidewalk surface, not the road.

Respond with ONLY a single number to ONE DECIMAL PLACE, in meters.
Do not include units, words, ranges, or explanation.
Example of a valid answer: 1.4
\end{verbatim}
}

\paragraph{Longitudinal (running) slope}
{\small
\begin{verbatim}
You are an expert analyzing the provided street view image.

Estimate the LONGITUDINAL SLOPE (the running grade) of the SIDEWALK
(the pedestrian footpath) along the walking direction, expressed in DEGREES.

Use a SIGNED value from the perspective of someone walking forward into the scene:
  positive (+) means UPHILL  (the path rises ahead),
  negative (-) means DOWNHILL (the path falls ahead),
  0 means flat/level.

Respond with ONLY a single signed number to ONE DECIMAL PLACE, in degrees.
Always include the sign. Do not include units, words, ranges, or explanation.
Example of a valid answer: +2.0
Another valid answer: -3.5
\end{verbatim}
}

\paragraph{Cross slope}
{\small
\begin{verbatim}
You are an expert analyzing the provided street view image.

Estimate the CROSS SLOPE of the SIDEWALK (the pedestrian footpath), expressed in DEGREES.

Cross slope means the tilt of the walking surface ACROSS the sidewalk,
perpendicular to the walking direction (not the uphill/downhill grade along the path).

Measure it from the RIGHT edge (starting point) to the LEFT edge (ending point),
as seen by someone walking forward into the scene:
  positive (+) means the surface RISES from right to left (the LEFT edge is HIGHER),
  negative (-) means the surface FALLS from right to left (the LEFT edge is LOWER),
  0 means the surface is level across.

Respond with ONLY a single signed number to ONE DECIMAL PLACE, in degrees.
Always include the sign. Do not include units, words, ranges, or explanation.
Example of a valid answer: +1.0
Another valid answer: -1.5
\end{verbatim}
}

\paragraph{Pavement condition}
{\small
\begin{verbatim}
You are an expert sidewalk inspector analyzing the provided street view image.

Evaluate the PAVEMENT SURFACE CONDITION of the SIDEWALK (the pedestrian footpath),
not the road. Judge only the pavement surface itself (cracks, deformation, broken or
missing pavers, potholes, heaving, settlement). Ignore temporary objects such as
parked bicycles, pedestrians, vehicles, or street furniture.

Assign exactly ONE grade using this rubric:

Grade A (Very Good): No plastic deformation or cracking; newly constructed
  or in like-new condition.
Grade B (Good): The surface is not entirely smooth but maintains a uniform
  condition overall.
Grade C (Fair): Passage by pedestrians and mobility-impaired users is possible,
  but resurfacing or similar measures may need to be considered depending on
  the degree of pavement deterioration.
Grade D (Poor): Deterioration severe enough to affect normal passage by
  pedestrians and mobility-impaired users. Hazards exist on 50% or more
  of the pavement.
Grade E (Very Poor): Normal passage is not possible. Hazards exist on 75%
  or more of the pavement.

Respond with ONLY the single capital letter of the grade: A, B, C, D, or E.
Do not include words, punctuation, or explanation.
Example of a valid answer: B
\end{verbatim}
}

\end{document}